%% file: kdd2023.tex
\documentclass[sigconf]{acmart}

\usepackage[T1]{fontenc}
\usepackage{algorithm}
\usepackage{algorithmicx}
\usepackage{algpseudocode}
\usepackage{multirow}
\usepackage{makecell}
\usepackage{textcomp}

\usepackage{amssymb}
\usepackage{amsmath}
\usepackage{bm}
\usepackage{bbding}
\usepackage{color}
\usepackage{pifont}
\usepackage{CJKutf8}
\usepackage{url}
\usepackage[normalem]{ulem}
\usepackage{subfigure}
\usepackage{paralist}
\useunder{\uline}{\ul}{}
\AtBeginDocument{%
  \providecommand\BibTeX{{%
    \normalfont B\kern-0.5em{\scshape i\kern-0.25em b}\kern-0.8em\TeX}}}

\setcopyright{acmcopyright}
\copyrightyear{2023}
\acmYear{2023}
\setcopyright{acmlicensed}\acmConference[KDD '23]{Proceedings of the 29th ACM SIGKDD Conference on Knowledge Discovery and Data Mining}{August 6--10, 2023}{Long Beach, CA, USA}
\acmBooktitle{Proceedings of the 29th ACM SIGKDD Conference on Knowledge Discovery and Data Mining (KDD '23), August 6--10, 2023, Long Beach, CA, USA}
\acmPrice{15.00}
\acmDOI{10.1145/3580305.3599891}
\acmISBN{979-8-4007-0103-0/23/08}

\begin{document}

\renewcommand{\algorithmicrequire}{\textbf{Input:}}  
\renewcommand{\algorithmicensure}{\textbf{Output:}} 

\title{QUERT: Continual Pre-training of Language Model for Query Understanding in Travel Domain Search}

\author{Jian Xie}
\authornote{Work done when interned at Alibaba Group.}
\affiliation{%
  \institution{Shanghai Key Laboratory of Data Science, School of Computer Science, Fudan University}
  \city{Shanghai}
  \country{China}
}
\email{jianx0321@gmail.com}

\author{Yidan Liang}
\affiliation{%
  \institution{Alibaba Group}
  \city{Hangzhou}
  \country{China}
}
\email{liangyidan.lyd@alibaba-inc.com}

\author{Jingping Liu}
\authornote{Jingping Liu and Yanghua Xiao are corresponding authors.}
\affiliation{%
  \institution{School of Information Science and Engineering, East China University of Science and Technology}
  \city{Shanghai}
  \country{China}
}
\email{jingpingliu@ecust.edu.cn}

\author{Yanghua Xiao}
\authornotemark[2]
\affiliation{%
  \institution{Shanghai Key Laboratory of Data Science, School of Computer Science, Fudan University}
  \city{Shanghai}
  \country{China}
}
\email{shawyh@fudan.edu.cn}

\author{Baohua Wu}
\affiliation{%
  \institution{Alibaba Group}
  \city{Hangzhou}
  \country{China}
}
\email{zhengmao.wbh@alibaba-inc.com}

\author{Shenghua Ni}
\affiliation{%
  \institution{Alibaba Group}
  \city{Hangzhou}
  \country{China}
}
\email{shenghua.nish@alibaba-inc.com}


\renewcommand{\shortauthors}{Xie and Liang, et al.}

\begin{abstract}
    \input{000abstract}
\end{abstract}




\maketitle

\section{Introduction}
\label{sec:intro}
\input{010introduction}

\section{Related Work}
\label{sec:related}
\input{020related}

\section{QUERT}
\label{sec:method}
\input{030method}

\section{Experiments}
\label{sec:experiment}
\input{040experiments}


\section{Conclusion and Future Work}
\label{sec:conclusion}
\input{060conclusion}

\section{Acknowledgements}
This work was supported by Alibaba Group through Alibaba Innovative Research Program, Shanghai Municipal Science and Technology Major Project (No.2021SHZDZX0103), Science and Technology Commission of Shanghai Municipality Grant (No. 22511105902), and Shanghai Sailing Program (No. 23YF1409400).
\bibliographystyle{ACM-Reference-Format}
\bibliography{kdd2023}

\end{document}

%% file: 000abstract.tex
In light of the success of the pre-trained language models (PLMs), continual pre-training of generic PLMs has been the paradigm of domain adaption.
In this paper, we propose \textbf{QUERT}, A Continual Pre-trained Language Model for \textbf{QUER}y Understanding in \textbf{T}ravel Domain Search.
QUERT is jointly trained on four tailored pre-training tasks to the characteristics of query in travel domain search: Geography-aware Mask Prediction, Geohash Code Prediction, User Click Behavior Learning, and Phrase and Token Order Prediction. 
Performance improvement of downstream tasks and ablation experiment demonstrate the effectiveness of our proposed pre-training tasks.
To be specific, the average performance of downstream tasks increases by 2.02\% and 30.93\% in supervised and unsupervised settings, respectively.
To check on the improvement of QUERT to online business, we deploy QUERT and perform A/B testing on Fliggy APP. 
The feedback results show that QUERT increases the Unique Click-Through Rate and Page Click-Through Rate by 0.89\% and 1.03\%  when applying QUERT as the encoder. Resources are available at \url{https://github.com/hsaest/QUERT}.

%% file: 010introduction.tex
\vspace{-1em}
\begin{figure}[tbh]
    \includegraphics[scale=0.17]{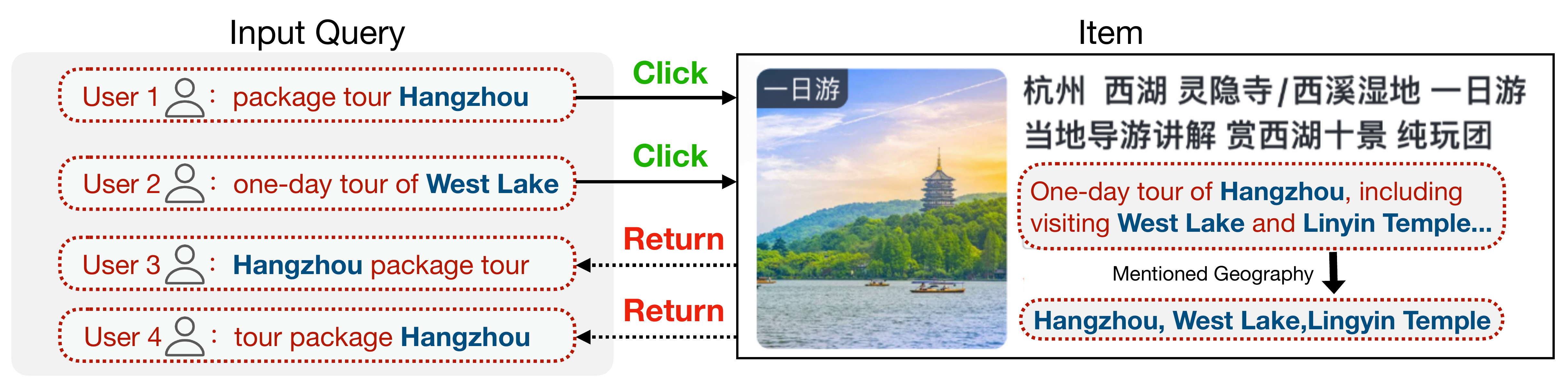}
    \caption{Examples of query and click item. ``Click'' in green means users click the item, and ``Return'' in red means the item is expected to return for the query.}
    \label{fig:examples}
\end{figure}
Pre-trained language models (PLMs) \cite{liu2019roberta,lan2019albert,yang2019xlnet,dong2019unified} have become the backbone models in the field of natural language processing (NLP) due to their superior representation ability.
Thus, PLMs are widely used in various domains and provide significant performance gains for downstream tasks (e.g., text classification \cite{garg2020bae} and information extraction \cite{liang2020bond}).

However, commonly used pre-trained models often perform poorly when directly transferred to specific domains (e.g., travel domain search).
This is caused by the mismatch between the corpora in the pre-training stage and the data in the target task.
In order to address this problem, previous work proposes to use domain corpora to continually pre-train domain-specific models, like BioBERT \cite{lee2020biobert} for bio-medicine, FinBERT \cite{araci2019finbert} for finance, and COMUS \cite{gong2022continual} for math problem understanding, etc.
With the domain adaption model, the performance of downstream tasks has been significantly improved, which validates the importance of continual pre-training on domain data.


\begin{CJK}{UTF8}{gkai}
In this paper, we focus on the continual pre-training for query understanding in travel domain search.
Travel domain search is the basis of an online travel APP, and there are many studies devoted to this aspect (e.g., named entity recognition \cite{cowan2015named} and hotel search ranking \cite{xu2022g2net}).
Nevertheless, to the best of our knowledge, there is no work focused on PLM in travel domain search.
In addition, due to the characteristics of the query in travel domain search, the previous pre-training tasks are not applicable.
We analyze the query in travel domain search and summarize three characteristics:
\textbf{(1) Geography Awareness.} Most user input queries consist of geographical location and intention (e.g., in ``package tour Hangzhou'', ``package tour'' is the intention, and ``Hangzhou'' is a tourist city in China). 
But classical MLMs fail to grasp the importance of geography because the random masking strategy treats all tokens equally.
Besides, the representation produced by MLM is based on contextual understanding, which means it can not reflect the real physical geography information.
\textbf{(2)  .} 
In the search logs, many different queries link to the same click item, which means they have similarity driven by user click behavior.
Therefore, PLM in travel domain search is expected to model such a potential similarity.
As shown in Figure \ref{fig:examples}, ``package tour Hangzhou'' and ``one-day tour of West Lake'' are two literally different queries, but they point to the same click item.
However, the conventional PLMs do not have such an ability because of lacking related pre-training tasks.
\textbf{(3) Robustness to Phrase and Token Order.} 
First, due to the user input habit, users might freely permutate the phrase order in a query.
For example, ``package tour | Hangzhou'' (``|'' denotes the phrase separator) might be entered as ``Hangzhou | package tour''.
Second, the token orders in the phrase might be transposed because of the user misinput (e.g., ``tour package | Hangzhou'').
We define the above phenomena as Phrase Permutation and Token Permutation, respectively.
In fact, the intentions in these two permutation cases are the same, and the returned results are expected to be the same.
Therefore, the model in travel domain is expected to be robust to phrase and token order.
However, due to lacking specific pre-training, conventional language models cannot understand the logical consistency in permutation.
\end{CJK}
According to the query characteristics in travel domain search, we propose QUERT to effectively learn query representations through four customized pre-training tasks.
Given a query, we introduce its click item as the additional information.
Specifically, to solve problem (1), we design a masking strategy called \textbf{Geography-aware Mask Prediction (Geo-MP)} to force the pre-trained model to pay more attention to the geographical location phrases.
In addition to semantic understanding, we introduce geohash in \textbf{Geohash Code Prediction (Geo-CP)} task to model real physical geographic information for the language model.
As for problem (2), in order to build a connection between the different queries linking to the same click item, we propose \textbf{User 
 Click Behavior Learning (UCBL)} to learn the potential similarity.
To solve problem (3), we propose \textbf{Phrase and Token Order Prediction (PTOP)}.
We shuffle the phrases and tokens to simulate the permutation.
QUERT is expected to predict the original phrases and tokens order of the shuffled query.
This task aims to enable QUERT to learn the logical consistency in permutation and be robust to phrase and token order.

Our contributions are summarized as follows:
\begin{inparaenum}[\it 1)]
    \item To the best of our knowledge, we are the first to explore continual pre-training for query understanding in the travel domain search.
    \item We propose four tailored pre-training tasks: Geography-aware Mask Prediction, Geohash Code Prediction, User Click Behavior Learning, and Phrase and Token Order Prediction.
    \item The experimental results on five downstream tasks related to travel domain search prove the effectiveness of our method. In particular, model performance improves by 2.02\% and 30.93\% under supervised and unsupervised settings, respectively. And the online A/B testing on Fliggy APP\footnote{Fliggy is an online travel agency in China. The official website is \url{www.fliggy.com}} demonstrates that QUERT improves the Unique Click-Through Rate and Page Click-Through Rate by 0.89\% and 1.03\% when applying QUERT as the feature encoder.
\end{inparaenum}

%% file: 020related.tex
The related work in this paper can be divided into three groups: open domain pre-trained language models, domain adaption pre-trained language models and query search.

\textbf{Open Domain PLMs.}
Based on the Transformer \cite{vaswani2017attention} architecture, BERT \cite{devlin-etal-2019-bert} has proven the effectiveness of large-scale corpora pre-training in natural language processing.
After that, other Transformer-based models \cite{lan2019albert,yang2019xlnet,brown2020language,lewis-etal-2020-bart} followed, either structural improvements or changes in pre-training tasks.
Instead of only considering representations based on token-level masking strategy, \citet{zhang2019ernie} adopt entity-level masking and phrase-level masking as the masking strategy.
Likewise, SpanBERT \cite{joshi2020spanbert} no longer masks individual tokens but continuous spans to enhance the model's text comprehension.

\textbf{Domain Adaption PLMs.}
\citet{gururangan2020don} point out that domain adaption PLMs provide large gains in domain tasks.
Based on the large corpora of financial texts, FinBERT \cite{araci2019finbert} gains outstanding performance in financial sentiment classification.
For Tweets, \citet{nguyen-etal-2020-bertweet} propose BERTweet to improve the performance on several Tweet NLP Tasks.
Besides, COVID-Twitter-BERT \cite{muller2020covid} achieves state-of-the-art (SOTA) performance on five classification tasks.
And the largest performance gap is in the COVID-19 classification task.
Excellent performance reflects the importance of domain adaption pre-training.

\textbf{Query Search.}
For query search, PROP \cite{ma2021prop} and B-PROP \cite{ma2021b} introduce representative words prediction task in the pre-training phase to model query and document.
Furthermore, it turns out that incorporating geographic information into a language model can make the model geo-sensitive.
For example, Baidu Maps apply ERNIE-GeoL \cite{huang2022ernie}, a language model pre-trained on whole word mask and geocoding prediction task, to improve the performance on geographical tasks.
In the Point of Interest (POI) search, \citet{liu2021geo} propose Geo-BERT to learn graph embeddings that simulate the POIs distribution.
However, due to the pre-defined single objective, the above work only considers a single characteristic of the query and thus cannot be generalized to other tasks in the search domain. 
Therefore, we propose custom tasks based on the common characteristics of the travel domain 
 search to improve the generalization of the model.

%% file: 030method.tex
\begin{figure*}[tb]
    \includegraphics[width=\linewidth]{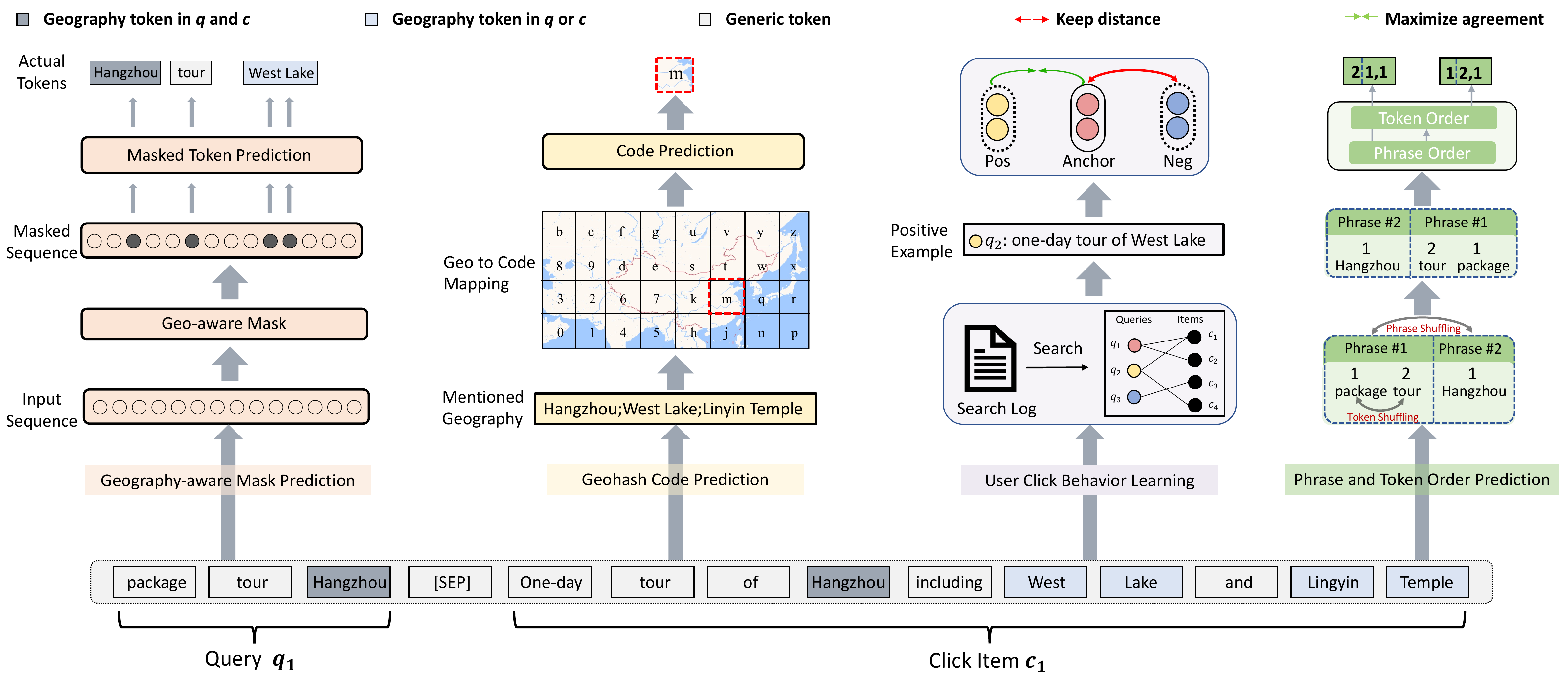}
    \caption{Our framework of QUERT. The Geohash Codes in the figure are only for presentation and do not represent the final code in our implementation.}
    \label{fig:model}
\end{figure*}





Given a query $q = [w_{q,1}, ..., w_{q,m}]$ and its click item title $c = [w_{c,1}, ..., w_{c,n}]$, where $m$ and $n$ are the numbers of tokens in $q$ and $c$, QUERT obtains the contextual representation for each token. 
As shown in Figure \ref{fig:model}, ``package tour Hangzhou'' is the query, and ``One-day tour of Hangzhou including visiting West Lake and Lingyin Temple'' is the related click item title. 
For convenience, we define the phrase as a group of tokens standing for a grammatical unit. 
For example, in the mentioned query,  ``package'' is a token, while ``package tour'' is a phrase. 
According to the characteristics of query in travel domain search, we design four pre-training tasks: \textbf{Geography-aware Mask Prediction} (Section \ref{subsec:geo-aware}), \textbf{Geohash Code Prediction} (Section \ref{subsec:geohash}), \textbf{User Click Behavior Learning} (Section \ref{subsec:cl}), and \textbf{Phrase and Token Order Prediction} (Section \ref{subsec:token-order}).


\subsection{Geography-aware Mask Prediction}
\label{subsec:geo-aware}
Geography-aware Mask Prediction (Geo-MP) aims to enable QUERT to be aware of geography information.
The reason is that we observe that most of the queries contain geography information in the travel domain search.
To verify this point, we randomly sample 1,000 queries from Fliggy APP and identify the geography information through inner geography dictionary mapping.
The statistical result shows queries containing geography information account for 65\%.
Therefore, the language models are expected to be good at the representation of geography-related queries.
However, the existing pre-trained language models (e.g., BERT) do not have this capability because of lacking specific pre-training tasks.
Thus, we propose the Geography-aware Mask Prediction to improve the situation.

Given query $q$ and its corresponding click item $c $, we use a special token ``[SEP]'' to combine them.
The reasons for incorporating click items are summarized as two points.
First, in our statistics, most of the query lengths are concentrated between 1 and 10. 
Short text length is not conducive to pre-training but increases the risk of weakening the model's representation ability (detailed experimental proofs are described in Section \ref{subsec:supervised-results}). 
Second, the click items are derived from real user behavior, which verifies the query and item are highly matched.
Besides, the length of the item title is appropriate (concentrating between 25 and 35).
Thus, the click item is the optimal solution to enrich query information.

Inspired by BERT, we apply masked language model (MLM) to predict the masked tokens, where the difference is we assign a higher probability to geography tokens.
To be specific, we use the NER tool\footnote{In this paper, we use AliNLP which is developed by Alibaba Damo Academy.} as a detector to identify geography phrases that appear in $q$ and $c$.
We design three mask strategies.
First, for geography phrases in both $q$ and $c$, we set a mask probability of 50\% for them. 
We consider that when common geography phrases are masked at the same time, the randomness of the prediction is excessive due to the lack of context.
Ideally, with this probability, for geography phrases in $q$ and $c$, when one of them is masked, the other is visible. This enables the model to infer the masked phrase from the visible one.
Second, we assign a mask probability of 30\% to the geography phrases that only appear in $q$ or $c$. 
Third, for the rest of the tokens, as with BERT, we mask 15\% of them.
The masked language model loss is given by:
\begin{equation}
\mathcal{L}_{Geo-MP} = -\sum_{i \in M_w} \log p\left(w_i \mid w_{\backslash M_w}\right),
\end{equation}
where $M_w$ is the positions of masked tokens.





\subsection{Geohash Code Prediction}
\label{subsec:geohash}
In addition to Geo-MP, we introduce Geohash Code Prediction (Geo-CP) to enhance the geography sensitivity of QUERT.
The reason is that in travel domain search, 
semantic understanding is not enough for downstream tasks (e.g., Query Rewriting and Query-POI Retrieval).
For example, in the Query-POI retrieval task, when the query is ``Hangzhou Tour'', ``West Lake'' is one of the potential recall because West Lake is a famous POI located in Hangzhou.
However, MLMs based on mask strategy can only understand geographical location from the semantic level and cannot capture the hierarchy or distance relationship between geographical locations.
So MLM may recall POIs mentioning ``Hangzhou'' while ignoring ``West Lake''.
Thus, QUERT is expected to have the ability to model real geographic locations (e.g., physical location distance and geographical hierarchy).
Therefore, we put forward Geo-CP.


Given click query and click items, the objective of Geo-CP is to predict the geohash code produced by the geohash algorithm.
The geohash algorithm divides geographic blocks into several grids and encodes them according to latitude and longitude.
The code is represented as a string, with precision controlled by the number of bits in the string.
Each bit represents a different granularity of geographic information.
And the adjacent grids share the same prefix.
We assume that the length of geohash code is $N$ bits. 
In order to encode the geography text, first, we locate the latitude and longitude of every geography entity in items.
Geography data uploaded by the service provider is considered to be of high confidence.
This is why we only consider the geography in item.
Second, we encode latitude and longitude into geohash code.
Finally, in order to get a unique encoding for each input, we process them separately according to the number of parsed geography units.
1) For item including no geography unit, we use $N$ bits special token ``*'' to stand for it.
2) For item including only one geography unit, we adopt its geohash code as the final geohash code.
3) For item including several different geography units, we adopt their longest prefix as the final geohash code. And the parts that are short of $N$ bits are filled with special tokens ``*''.
In terms of model architecture, we use N independent multi-layer perceptrons (MLPs) to predict bits at different positions.
In other words, each MLP has its own granularity prediction capability.
The Geo-CP loss is defined as:
\begin{equation}
\mathcal{L}_{Geo-CP}=-\frac{1}{N} \sum_{i=1}^L y_i \log p_i,
\end{equation}
where $L$ is the number of potential characters in one bit.



\subsection{User Click Behavior Learning}
\label{subsec:cl}
In the search logs, we observe that literally different queries may point to the same click item.
For instance, ``package tour Hangzhou'' and ``one-day tour of West Lake'' are both related to the item ``One-day tour of Hangzhou including visiting West Lake and Lingyin Temple''.
These two queries are not literally similar, but they have an implicit query similarity driven by user click behavior.
Conventional MLMs are unable to model this implicit similarity because of lacking specific pre-training.
Therefore, we propose User Click Behavior Learning (UCBL) based on contrastive learning.

Formally, given a query $q_{i}$ and its click item $c_{i}$, according to the click rate, we select the top K from the queries linking to $c_{i}$ and combine them into a group $G= \{q_{i1},q_{i2},...,q_{iK}\}$.
Then, in order to guarantee the diversity, we randomly choose one from $G$ as the positive example $q_{i}^{pos}$ of $q_{i}$.
As for other queries $q_{j}$ in batch, we regard them and the corresponding positive examples $q^{pos}_{j}$ as the negative examples for $q_{i}$.
Feeding $q_{i}$ and $q^{pos}_{i}$ into the encoder (i.e., QUERT), we adopt the ``[CLS]'' embedding to represent the input.
The embeddings are represented as $\mathcal{R}_{i}$ and $\mathcal{R}^{pos}_{i}$ respectively.
The optimizing objective can be expressed as following:
\begin{equation}
\mathcal{L}_{UCBL}=-\log \frac{\exp({\mathit{sim}(\mathbf{\mathcal{R}_\mathit{i}}, \mathit{\mathcal{R}^{pos}_{\mathit{i}}} )/ \tau})}{\sum_{j=1}^\mathit{N} \exp({\mathit{sim}(\mathit{\mathcal{R}_{j}}, \mathit{\mathcal{R}^{pos}_{j}}) / \tau})},
\end{equation}
The sim(·) function which we use is cosine similarity, and $\tau$ is used for smoothing the distribution.




\subsection{Phrase and Token Order Prediction}
\label{subsec:token-order}
\begin{CJK}{UTF8}{gkai}
Phrase and Token Order Prediction (PTOP) is designed to enable QUERT to learn the logical consistency in permutation, thus being robust to phrase and token order.
According to our observation, the permutation query in the travel domain has the following two types.
1) Phrase Permutation. Queries are presented in different forms because of users' personalized input habits.
For example, the query ``package tour Hangzhou'' would be entered as ``Hangzhou package tour''.
This would lead to differences in the returned results due to the change of query.
In essence, for the same intention queries, the returned results are expected to be the same.
2) Tokens in Phrase Permutation.
User misinput causes the token permutation in the query.
Taking the same case ``package tour Hangzhou'' for example, wrong results will be returned if the user enters ``tour package Hangzhou''.
In our statistics, the inappropriate returned results caused by token permutation account for 5.3\% in 5000 randomly selected bad cases.
However, conventional PLMs are not good at modeling such logical consistency in permutation.
To this end, we propose Phrase and Token Order Prediction in this paper.

Given the original query ``package tour Hangzhou'', QUERT is expected to predict every token's corresponding phrase order and token order after permutation.
We pre-define the phrase order for every token as $(1,1,2)$.
``package'' and ``tour'' are both in the first phrase, so their phrase orders are both ``1''.
And the token order for every token is $(1,2,1)$.
Note that every token order is limited to the phrase to which the token belongs.
In other words, the maximum token order is no greater than the length of its corresponding phrase.
Specifically, in the phrase ``package tour'', ``package'' is the first token, and ``tour'' is the last token, so their token order is ``1'' and ``2'' respectively.
And ``Hangzhou'' belongs to the second phrase, so its token order is back to ``1''.

To simulate the permutation, first, we randomly shuffle phrases in a query.
So the ``package tour Hangzhou'' would be `Hangzhou package tour'', and the ground truth phrase order is defined as $y=(2,1,1)$.
Second, we randomly select phrases with a specific probability and shuffle the tokens in them.
Under this setting, we assume the selected phrase is ``package tour''.
After shuffling, the phrase would be ``tour package''.
And the final shuffled query is ``Hangzhou tour package''.
Therefore, in this case, the ground truth of the token order is $y=(1,2,1)$.
Lastly, the output order is computed by the tokens layer and phrase layer, respectively.
Since token order is predicted based on the phrase, we design the tokens layer following phrase layer:
\begin{equation}
\begin{aligned}
\begin{split}
\left[r_1, \ldots, r_i, \ldots, r_m\right]&=QUERT\left(\left[w_{q,1}, w_{q,2}, \ldots, w_{q,m}\right]\right),
\\
p_{i}^{\alpha}&=\operatorname{Softmax}(\operatorname{MLP_{phrase}}(r_{i})),
\\
p_{i}^{\beta}&=\operatorname{Softmax}(\operatorname{MLP_{token}}(\operatorname{MLP_{phrase}}(r_{i}))).      
\end{split}
\end{aligned}
\end{equation}
The training formulation is given by:
\begin{equation}
\small
\mathcal{L}_{TPOP}=-\frac{1}{m} \sum_{i=1}^m \left(\sum_{c=1}^{Q} y_{i,c}^{\alpha} \log \left(p_{i,c}^{\alpha}\right)+ \sum_{d=1}^{R} y_{i,d}^{\beta} \log \left(p_{i,d}^{\beta}\right)\right)
\end{equation}
where $Q$ is the pre-defined maximum number of phrases, and $R$ is the pre-defined maximum token number in phrase. 
Note that we only predict the order of the tokens and phrases in query.
\end{CJK}

\subsection{Loss Function}
\label{subsec:loss-function}
In order to enable QUERT to integrate the above capabilities, we use joint training to combine four tasks. 
We consider each task to be equally important, so no additional weight is assigned.
And the overall training loss function is defined as:
\begin{equation}
\mathcal{L}= \mathcal{L}_{Geo-MP} + \mathcal{L}_{Geo-CP} + \mathcal{L}_{UCBL}+\mathcal{L}_{TPOP},
\end{equation}
The pre-training target is to minimize the $\mathcal{L}$.

%% file: 040experiments.tex
In this section, we demonstrate the effectiveness of our model through extensive offline experiments. 
Then, we deploy the model on Fliggy APP to test its performance in real online scenarios.
\subsection{Experimental Setup}
\label{subsec:experimental-setup}
\subsubsection{Pre-training corpora}
\label{subsubsec:corpora}

We collect data from the past three years of Fliggy's real online business scenarios. 
The data consists of user search query $q$, click items $c$, and their confidence scores (i.e., unique visitor click hits $UV_{C}$ and unique visitor payment hits $UV_{P}$).
Since the original data from online business contains numerous noise, we rank the data according to the weighted score 
$S=UV_C+10 \times UV_P$. 
To ensure the quality of the data, we select the top 5 million highest weighted score pairs $P=(q,c)$ and discard others. 

\subsubsection{Downstream Tasks}
\label{subsubsec:downstran-task}
The downstream tasks are described as follows.

\textbf{\emph{Query Rewriting (QR).}}
QR aims to reduce the expression gap between user queries and items.
In other words, QR rewrites an unparsed query into a parsed query which is stored in the database.
To build a high confidence dataset, first, we collect unparsed user input queries and the corresponding click items from search logs. 
Then we select the parsed query linked to the same click item as the candidate samples.
Finally, to ensure the quality of the dataset, we invite people to annotate the candidate samples and filter those that are marked as untrustworthy.

For evaluation, we select 181,405 standard queries as candidates.
We adopt accuracy (Acc) and  hit rate (Hits@K) as evaluation metrics. 
Hits@K means the success rate of finding the ground truth in top K candidates.
In line with the actual business, we set K to 20 in our experiments.
In order to retrieve, we feed the query into an encoder (e.g., BERT, RoBERTa, and QUERT) and gain its embedding.
We calculate the cosine similarity between the tested query and all candidates.
The top K highest cosine similarity sample would be used for evaluation.
In the supervised setting, we use contrastive learning to fine-tune the similarity.
Specifically, the ground truth is regarded as a positive example, and negative examples are randomly sampled in the training batch.

\textbf{\emph{Query-POI Retrieval (QPR).}}
Given a query, QPR aims to provide several POI recommendations to improve users' search experience.
In order to construct the dataset, for every query, we select the top 20 click rate POIs as ground truths.
All data is manually verified again.

For evaluation, we select 201,184 POIs as candidates for retrieval.
Accuracy (Acc) and  hit rate (Hits@K) are adopted as evaluation metrics.
And K is set to 20.
The implementation detail of training is the same as QR.

\textbf{\emph{Query Intention Classification (QIC).}}
For precise recommendation of relevant items, QIC aims to predict the user's intention based on the input query.
Based on business practices, we pre-define 20 categories of intentions.
For dataset building, first, we extract the item category with the highest click rate corresponding to the query as the intention pseudo-label.
For these data, we perform additional human checks and construct the dataset.

We regard this task as a sentence multi-class classification task. 
In terms of evaluation metrics, we adopt the widely used precision (P), recall (R) and F1.

\textbf{\emph{Query Destination Matching (QDM).}}
Given a query and candidate city, QDM aims to determine whether the city is the intended destination of the query.
This task adds constraints to the search recall module, avoiding irrelevant recall results, thereby improving the accuracy and relevance of the entire search engine.

Similar to the intention classification task, we select the location of the click item with the top K highest click rate as the destination label.
We invite people to do a further evaluation of the data and build the dataset.
Given a pair consisting of a query and a city, the model performs binary classification of the pair.
We use precision, recall and F1 to evaluate the quality.

\textbf{\emph{Query Error Detection (QED).}}
QED identifies specific token errors within queries, such as typos or incorrect order, rather than classifying the entire query as wrong. 
Improved error detection through QED enables more accurate search results.
We collect wrong queries from real data, including redundant tokens, transpositions of token order, and typos, to build the final dataset.
There are four kinds of labels in total. 
Labels ``0'' to ``3'' signify no error, a typo, token transposition, and redundant token respectively. 
We regard QED as a sequence-labeling task, so the model is expected to predict the specific error for each token.
Precision, recall and F1 of token-level are used as evaluation metrics.

The detailed data scale and evaluation metrics of five downstream tasks are presented in Table \ref{tab:task-statistic}. 

\begin{table}[tb]
\small
\caption{Statistics of downstream tasks dataset. ``\#'' indicates the number of samples.}
\begin{tabular}{lcccc}
\hline
\multicolumn{1}{c}{\multirow{2}{*}{TASK}} & \multicolumn{3}{c}{\#} & \multirow{2}{*}{Metric} \\ \cline{2-4}
\multicolumn{1}{c}{}                      & Train  & Dev   & Test  &                         \\ \hline
Query Rewriting                       & 20,000  & 2,500  & 2,500  & Acc, Hits@20                   \\
Query-POI Retrieval                        & 26,854  & 3,836 & 2,947 & Acc, Hits@20                  \\
Query Intention Classification            & 54,003  & 6,712  & 6,709  & P, R, F1                  \\
Query Destination Matching              & 67,241  & 7,982  & 8,443  & P, R, F1                  \\
Query Error Detection                     & 48,333  & 6,890  & 6,794  & P, R, F1                  \\ \hline
\end{tabular}
\label{tab:task-statistic}
\vspace{-1.5em}
\end{table}

\begin{table*}[tb]
\small
\caption{Supervised performance comparison of different setting models on downstream tasks. All results are based on the average of five replicates. BERT$_{q}$ is BERT continually pre-trained on only query corpora, and BERT$_{q+c}$ is continually pre-trained on query and item. The subscript of QUERT indicates the backbone model. The best results are marked in bold and the second best are underlined.}
\begin{tabular}{lcccccccccccccc}
\toprule
\multirow{2}{*}{Models} & \multicolumn{2}{c}{QR}             & \multicolumn{2}{c}{QPR} & \multicolumn{3}{c}{QIC} & \multicolumn{3}{c}{QDM} & \multicolumn{3}{c}{QED} & \multirow{2}{*}{Average} \\ \cmidrule(l){2-3} \cmidrule(l){4-5} \cmidrule(l){6-8} \cmidrule(l){9-11} \cmidrule(l){12-14}
                        & Acc  & \multicolumn{1}{l}{Hits@20} & Acc  & Hits@20 & P      & R      & F1    & P      & R      & F1    & P      & R      & F1    &                          \\ \midrule
BERT \cite{devlin-etal-2019-bert}                   & 47.94 & 92.53                       & 59.71 & 79.97   & 41.70  & 35.08  & 35.56 & 90.18  & 88.36  & 89.20 & 89.55  & 87.39  & 88.35 & 71.19                    \\
RoBERTa \cite{liu2019roberta}                & 47.93 & 92.72                       & 59.43 & 80.37   & 42.07  & 33.55  & 34.56 & 90.21  & 88.13  & 89.07 & 89.96  & 86.94  & 88.27 & 71.02                    \\
ERNIE \cite{sun2019ernie}                  & 47.52 & 92.81                      & 60.54 & 82.23   & 41.84  & 35.54  & 36.06 & 89.22  & 88.78  & 88.98 & 89.57  & 87.22  & 88.27 & 71.43                    \\
StructBERT \cite{wang2019structbert}             & 47.51 & 92.35                       & 60.00 & 81.34   & 41.90  & 35.44  & 35.87 & 90.00  & 88.90  & 89.42 & 89.79  & 85.75  & 87.63 & 71.22                    \\
Mengzi  \cite{zhang2021mengzi}                & 47.87 & 92.05                       & 59.60 & 79.57   & 40.98  & 34.46  & 34.60 & 88.89  & 87.08  & 87.92 & 89.95  & \uline{87.48}  & 88.57 & 70.69                    \\ \midrule
BERT$_{q}$          & 47.80 & 92.96                       & 59.76 & 79.68   & 41.09  & 35.00  & 35.54 & 89.55  & 86.97  & 88.10 & 89.09  & 86.19  & 87.42 & 70.70                    \\
BERT$_{q+c}$     & 48.47 & \uline{93.04}                       & 59.94 & 80.68   & $\bf{45.67}$  & 35.90  & 36.93 & 89.24  & \uline{90.11}  & 89.63 & 90.17  & 87.19  & 88.53 & 71.96                   \\
QUERT$_{BERT}$          & \uline{48.66} & $\bf{93.08}$                       & $\bf{61.26}$ & $\bf{82.49}$   & \uline{45.44}  & $\bf{39.20}$  & $\bf{39.58}$ & $\bf91.78$  & \uline{90.11}  & $\bf{90.88}$ & \uline{91.25}  & $\bf88.30$  & $\bf89.64$ & $\bf73.21$                    \\
QUERT$_{ERNIE}$         & $\bf{48.67}$      & 92.99         & \uline{60.98}      & \uline{82.25}        & 43.24       &  \uline{38.41}      &   \uline{38.30}    &   \uline{90.77}     & $\bf90.96$       & \uline{90.87}      & $\bf91.86$       &   87.00     &  \uline{89.20}     &   \uline{72.73}                       \\ \bottomrule
\end{tabular}
\label{tab:donwstream-result}
\end{table*}

\begin{table}[tb]
\caption{Unsupervised performance comparison of different setting models on downstream tasks.}
\begin{tabular}{lccccl}
\toprule
\multirow{2}{*}{Models} & \multicolumn{2}{c}{QR}                      & \multicolumn{2}{c}{QPR}                     & \multirow{2}{*}{Average}  \\ \cmidrule(l){2-3} \cmidrule(l){4-5}
                        & Acc                 & Hits@20              & Acc                 & Hits@20              &                           \\ \midrule
BERT                    & 17.89                & 40.59                & 21.95                & 33.70                & \multicolumn{1}{c}{28.53} \\
ERNIE                   & 14.49                & 32.27                & 17.48                & 25.45                & \multicolumn{1}{c}{22.42} \\
QUERT$_{BERT}$        & $\bf{41.75}$                & $\bf{85.79}$                & \uline{44.15}                & \uline{63.76}                & \multicolumn{1}{c}{$\bf{59.46}$} \\
QUERT$_{ERNIE}$        & \uline{39.79} & \uline{85.03} & $\bf{44.32}$ & $\bf{64.54}$  & \multicolumn{1}{c}{\uline{58.42}}                          \\ \bottomrule
\end{tabular}
\label{tab:unsupervise-results}
\end{table}

\begin{table*}[tbh]
\caption{Ablation results of QUERT$_{BERT}$. We repeat five times and report the average scores. }
\small
\begin{tabular}{lcccccccccccccc}
\toprule
\multirow{2}{*}{Models} & \multicolumn{2}{c}{QR}             & \multicolumn{2}{c}{QPR} & \multicolumn{3}{c}{QIC} & \multicolumn{3}{c}{QDM} & \multicolumn{3}{c}{QED} & \multirow{2}{*}{Average} \\ \cline{2-14}
                        & Acc  & \multicolumn{1}{l}{Hits@20} & Acc      & Hits@20    & P      & R      & F1    & P      & R      & F1    & P      & R      & F1    &                          \\ \midrule
QUERT$_{BERT}$          & $\bf{48.66}$ & $\bf{93.08}$                       & 61.26     & $\bf{82.49}$      & 45.44  & $\bf{39.20}$  & $\bf{39.58}$ & $\bf{91.78}$  & 90.11  & $\bf{90.88}$ & $\bf{91.25}$  & $\bf{88.30}$  & $\bf{89.64}$ & $\bf{73.21}$                    \\
 \quad - w/o Geo-MP            & 48.05 & 93.05                       & 60.81     & 81.94      & $\bf{45.89}$  & 36.93  & 38.44 & 90.82  & 90.58  & 90.68 & 90.91  & 87.16  & 88.84 & 72.62                    \\
 \quad - w/o Geo-CP            & 47.61 & 92.31                       & 61.32     & 82.35      & 43.02  & 37.00  & 37.35 & 89.65  & $\bf{91.08}$  & 90.32 & 90.57  & 87.37  & 88.80 & 72.21                    \\
\quad - w/o UCBL               & 48.08 & 92.87                       & 60.70     & 82.44      & 44.00  & 36.50  & 37.01 & 90.04  & 88.72  & 89.33 & 90.66  & 86.57  & 88.39 & 71.95                  \\ 
 \quad - w/o PTOP              & 48.61 & 92.95                       & $\bf{61.42}$     & 82.22      & 44.95  & 37.23  & 38.23 & 90.92  & 90.77  & 90.84 & 88.73  & 85.60  & 86.99 & 72.25                    \\
\bottomrule
\end{tabular}
\label{tab:ablation-study}
\end{table*}

\subsubsection{Baselines}
We select several pre-trained models widely used in Chinese NLP tasks as the baselines for comparison with QUERT.
\begin{compactitem}
    \item \textbf{BERT}\cite{devlin-etal-2019-bert} is a strong PLM based on Transformer architecture. We use the Chinese version released by Google.
    \item \textbf{RoBERTa}\cite{liu2019roberta} is a PLM whose architecture is the same as BERT. We use the Chinese version RoBERTa-wwm \cite{cui2021pre}.
    \item \textbf{ERNIE}\cite{sun2019ernie} is also a Transformer-based PLM. In addition to the token-level mask, it introduces the entity-level and phrase-level masking strategies.
    \item \textbf{StructBERT}\cite{wang2019structbert} is the variant of BERT. It adds a new word structural objective in the training stage to force the model to reconstruct the right order of sequence, which is similar to our PTOP task.
    \item \textbf{Mengzi}\cite{zhang2021mengzi} is also the variant of BERT, which is dedicated to being lightweight but powerful. Mengzi gains outstanding performance on multiple Chinese NLP tasks.
\end{compactitem}

\subsubsection{Implementation Details}
\label{subsec:impl-details}
Our implementation is based on Transformers framework \cite{wolf-etal-2020-transformers} and Pytorch \cite{NEURIPS2019_bdbca288}. 
To verify the effectiveness of our proposed pre-training task on different model architectures, we apply BERT \cite{devlin-etal-2019-bert} and ERNIE \cite{zhang2019ernie} as backbone models to implement continual pre-training. 
We adopt AdamW \cite{loshchilov2018decoupled} as the optimizer and set the initial learning rate to 5e-5 for pre-training and 1e-5 for fine-tuning. 
With steps increasing, we decrease the learning rate linearly.  
The size of geohash code is 6 bits.
And we adopt Base32 as the numeral system in geohash coding.
The temperature $\tau$ of contrastive learning is set to 0.1.
And the shuffling probability in TPOP is 0.15.
The training takes about 72 hours for 386460 steps on 8 Tesla V100 GPUs with 16 batch size per device.

\subsection{Offline Results}
\label{subsec:main-results}

\subsubsection{Supervised results}
\label{subsec:supervised-results}
We compare QUERT with baseline models on five downstream tasks in the supervised setting (the model is fine-tuned on train set).
Table \ref{tab:donwstream-result} shows the results.

First, \textbf{QUERT achieves SOTA results in all downstream tasks}, which demonstrates the effectiveness of our proposed pre-training tasks. 
Specifically, compared with BERT, $QUERT_{BERT}$ improves the average performance by 2.02\%.
Furthermore, we find that the advantage of QUERT is more pronounced on difficult tasks.
In detail, QUERT has a huge performance advantage on QIC, which has an average improvement is nearly 4\%.

Second, \textbf{the corpora composed of single query information bring negative effects.}
We test directly continual pre-training on raw query corpora with the masking strategy used in BERT \cite{devlin-etal-2019-bert}, which is presented as BERT$_q$ in Table \ref{tab:donwstream-result}.
Experimental results show that BERT$_q$ brings the risk of negative effects. 
Specifically, BERT$_q$ achieves an average score of 70.90\% which is even lower than the original BERT.
We analyze that the model may not be able to learn knowledge representation from the short query but weaken the text understanding ability.
The results demonstrate that the regular pre-training task cannot be directly generalized to the pre-trained model focusing on the query.
This verifies that our proposed pre-training tasks are more applicable to travel domain.

Third, \textbf{the integration of click items information improves the representation ability of the model}.
We concatenate the title text information of items into the query to construct the new corpora.
Compared to BERT$_q$, the performance 
 of BERT$_{q+c}$ improves nearly 1.3\%. 
This verifies our conjecture that in the pre-training stage, the model is informed of the items' information with high confidence, which is helpful for the model to acquire more knowledge of the query.

\subsubsection{Unsupervised results}
\label{subsec:unsupervised-results}
We compare QUERT with two backbone models in the unsupervised setting (without fine-tuning).
Table \ref{tab:unsupervise-results} reports the unsupervised results.
We select QR and QPR as unsupervised test tasks because they can directly obtain results by calculating the embedding similarity.
We obtain the prediction by calculating the cosine similarity between the embeddings that are gained from the last layer of hidden states.
 ``[CLS]'' token is used for representing the whole query.
 
From the table, we observe that QUERT has significant performance advantages in unsupervised setting.
Both QUERT$_{BERT}$ and QUERT$_{ERNIE}$ significantly outperform the backbone model.
For QR, our proposed QUERT$_{BERT}$ outperforms the baseline BERT by 45.20\% on Hits@20.
As for QPR, QUERT$_{BERT}$ and QUERT$_{ERNIE}$ significantly outperform the two baselines, with a maximum performance gap of 39.09\% on Hits@20.
In terms of average scores, QUERT$_{BERT}$ achieves a 30.93\% performance improvement over BERT.
We analyze that the reason for the large performance gap is that QUERT is more potent in query understanding for travel domain search.
In unsupervised setting, the tailored pre-training tasks empower QUERT to give better query representation.
\vspace{-1em}

\subsubsection{Ablation Studies}
\label{subsubsec:Ablation Studies}
To check whether the customized pre-training tasks can effectively improve the performance of downstream tasks, we perform a series of ablation experiments. 
Specifically, we remove tasks one at a time to evaluate the impact of their presence or absence on performance.
Table \ref{tab:ablation-study} reports the results.

First, the removal of any tasks leads to performance loss.
The elimination of every component, i.e., Geo-MP, Geo-CP, UCBL, and PTOP, causes the 0.59\%, 1.00\%, 1.26\%, and 0.96\% drop in the score.

Second, we find that the removal of Geo-MP (i.e., degenerate to the original masking strategy) and Geo-CP results in performance degradation on all tasks.
This reveals that these two geography awareness pre-training tasks make an effective contribution to improving the ability of pre-trained model to perceive geography.


Third, UCBL plays the most important role in sentence-level tasks (i.e., QIC and QDM).
On the one hand, the construction of similarities in user behavior enables QUERT to understand queries better.
On the other hand, in the study of negative examples, the differentiation of sentence-level embedding representation is enlarged, which is beneficial to model recognition in the prediction.

Finally, the removal of PTOP leads to the F1 score on QED (86.99\%) being lower than the original BERT (88.35\%).
We guess that other tasks may introduce additional logical bias, which results in a performance penalty for the model on QED. 
However, the introduction of PTOP allows QUERT to refactor its understanding of the logical consistency of query, resulting in significant performance improvements.
In addition, we find that StructBERT, which aims at reconstructing order in MLM, did not achieve superior performance in QED.
This verifies that our proposed PTOP is more suitable for the real downstream task of travel domain search.

In conclusion, the results verify that our designed pre-training task in a tailored way endows the pre-trained model with powerful query representation capabilities in travel domain search.

\begin{figure}[t]
    \includegraphics[scale=0.155]{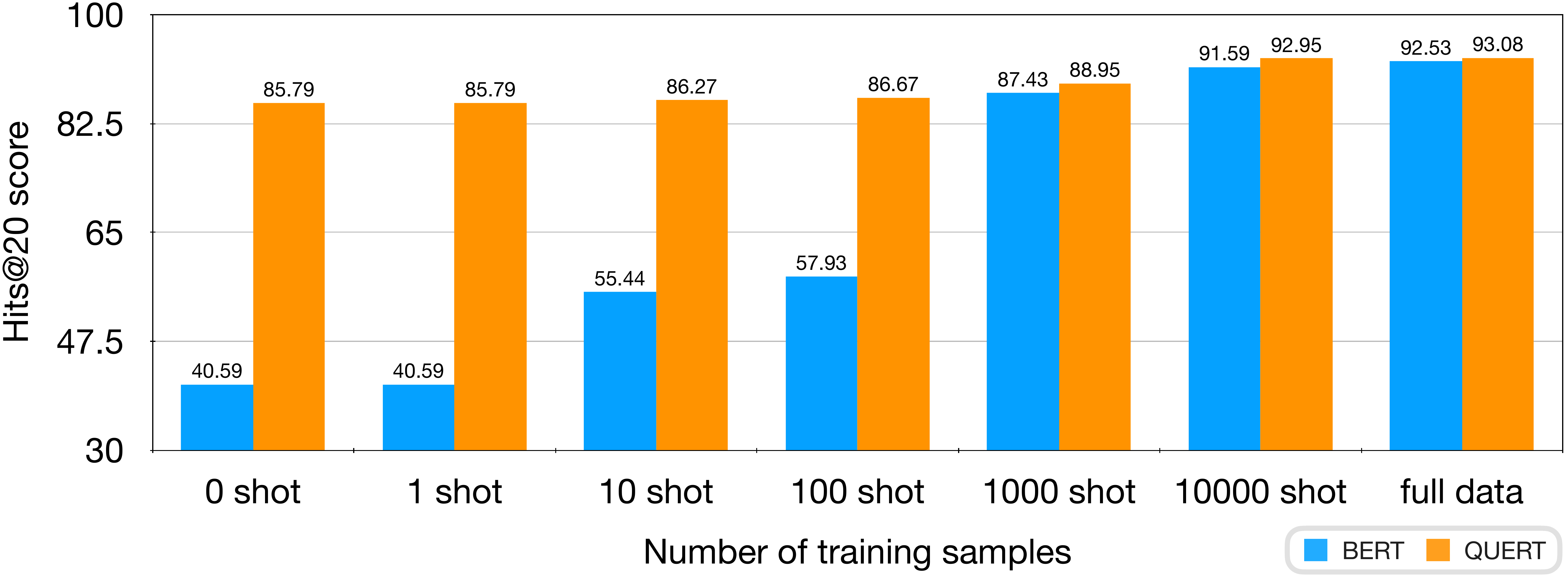}
    \caption{The comparison of few-shot learning on QR between BERT and  QUERT$_{BERT}$. We use Hits@20 as the  evaluation metric. }
    \label{fig:few-shot}
\end{figure}

\subsubsection{Few-shot Learning}
\label{subsubsec:few-shot}
In order to test the performance of QUERT in the data scarcity scenario, we conduct the few-shot learning experiment.
We choose QR as the experimental object because it is the most dependent on annotated data in real business, which can be verified in Table \ref{tab:unsupervise-results} (only 17.89\% Acc score in zero-shot).
Figure \ref{fig:few-shot} reports the results.
We conclude that:
1) QUERT can achieve better results than BERT under the same training data size.
2) When the amount of labeled data is no more than 100, the performance gap between QUERT and BERT is noticeable, ranging from 28.74\% to 45.20\%.
3) BERT does not surpass the zero-shot performance of QUERT until the training sample number reaches 1000.
The experimental results show that, compared with BERT, QUERT is competent for downstream tasks under sparse data scenarios, even under the extreme conditions of zero-shot and one-shot.

\subsection{Task Study}
\label{subsec:task-study}
In this section, we continually pre-train BERT with a single task to verify the true effect of each task.

\subsubsection{TASK 1: Geo-MP}
\label{subsubsec:Geo-MP}
In order to verify the effectiveness of Geo-MP, we evaluate the sensitivity to the geography of BERT and BERT+Geo-MP.
We randomly collect 500 queries from Fliggy APP. 
First, for geography token in query, we use a special token ``[mask]'' to hide it.
Then we let the model predict these masked tokens.
We adopt the hit rate score (denoted as Hits@K) and mean rank (i.e., MR and MRR) as the metric.
As shown in Table \ref{tab:geo-mask-report}, the best score of BERT is only 2.2\% on Hits@5.
In contrast, BERT+Geo-MP gains 13.2\% on Hits@5, lower MR and higher MR.

In order to further analyze the advantages of Geo-MP in location prediction, we present the cases in Table \ref{tab:geo-mp-case}.
In case 1 and 2, BERT+Geo-MP predict the right masked token, but BERT output the nonstandard city name or phrase.
Besides, in case 3, although BERT+Geo-MP gives a wrong geography phrase, it still outputs a reasonable prediction.
In fact, ``Ling yin Temple'' is exactly a POI located in Hangzhou, and it makes sense in the given context.
As another bad case, case 4 shows that the output of BERT+Geo-MP is closer to the answer.

The results verify that classical MLMs are indifferent to geography information.
By contrast, Geo-MP enables BERT to be more sensitive to geography.

\begin{table}[t]
\caption{Geography sensitivity comparison for BERT and BERT+Geo-MP. Hits@K means the success rate of finding the ground truth in K candidates. MR is the mean rank. MRR is the mean reciprocal rank.}
\begin{tabular}{llllcc}
\hline
      & Hits@1 & Hits@3 & Hits@5 & MR   & MRR  \\ \hline
BERT  & 1.2\%  & 1.8\%  & 2.2\%  & 1.91 & 0.71 \\
BERT+Geo-MP & \textbf{9.8\%} & \textbf{12.8\%} & \textbf{13.2\%} & \textbf{1.41} & \textbf{0.85} \\ \hline
\end{tabular}
\label{tab:geo-mask-report}
\end{table}

\begin{CJK}{UTF8}{gkai}
\begin{table}[t]
\tiny
\caption{Geography mask cases. ``*'' is the placeholder of ``MASK'' token. ``\ding{55}'' indicates the prediction result is not a phrase in Chinese.}
\begin{tabular}{|l|l|l|l|l|}
\hline
No & Text                                          & BERT+Geo-MP         & BERT                      & Answer            \\ \hline
1  & {[}**{]}天涯海角景区                                & 三亚                  & 北州                        & 三亚                \\ \hline
   & {[}**{]}Tianya Haijiao                        & Sanya               & \ding{55}    & Sanya             \\ \hline
2  & 北京{[}***{]}长城                                 & 八达岭                 & 的长山                       & 八达岭               \\ \hline
   & The Great Wall of  {[}***{]} Section, Beijing & Ba da ling          & \ding{55}   & Ba da ling        \\ \hline
3  & 杭州{[}***{]}门票                                 & 灵隐寺                 & 天物园                       & 岳王庙               \\ \hline
   & The ticket of {[}***{]}, Hangzhou             & Ling yin Temple     & \ding{55}  & Yue Fei Temple    \\ \hline
4  & 扬州{[}***{]}二十四桥                               & 瘦西街                 & 市桥第                       & 瘦西湖               \\ \hline
   & Yangzhou {[}***{]} Twenty-four Bridges        & Slender West Street & \ding{55}  & Slender West Lake \\ \hline
\end{tabular}
\label{tab:geo-mp-case}
\end{table}
\end{CJK}

\subsubsection{TASK 2: Geo-CP}
\label{subsubsec:Geo-CP}

To verify the physical geography representation ability of Geo-CP, we select 500 hot POIs in 10 popular cities and visualize their embeddings through t-SNE\cite{van2008visualizing}.
We use the embedding of ``[CLS]'' token to represent the POI. 

\begin{figure}[t]
	\subfigtopskip=2pt 
	\subfigure[BERT]{
		\label{subfig-type}
		\includegraphics[width=4cm,height=4cm]{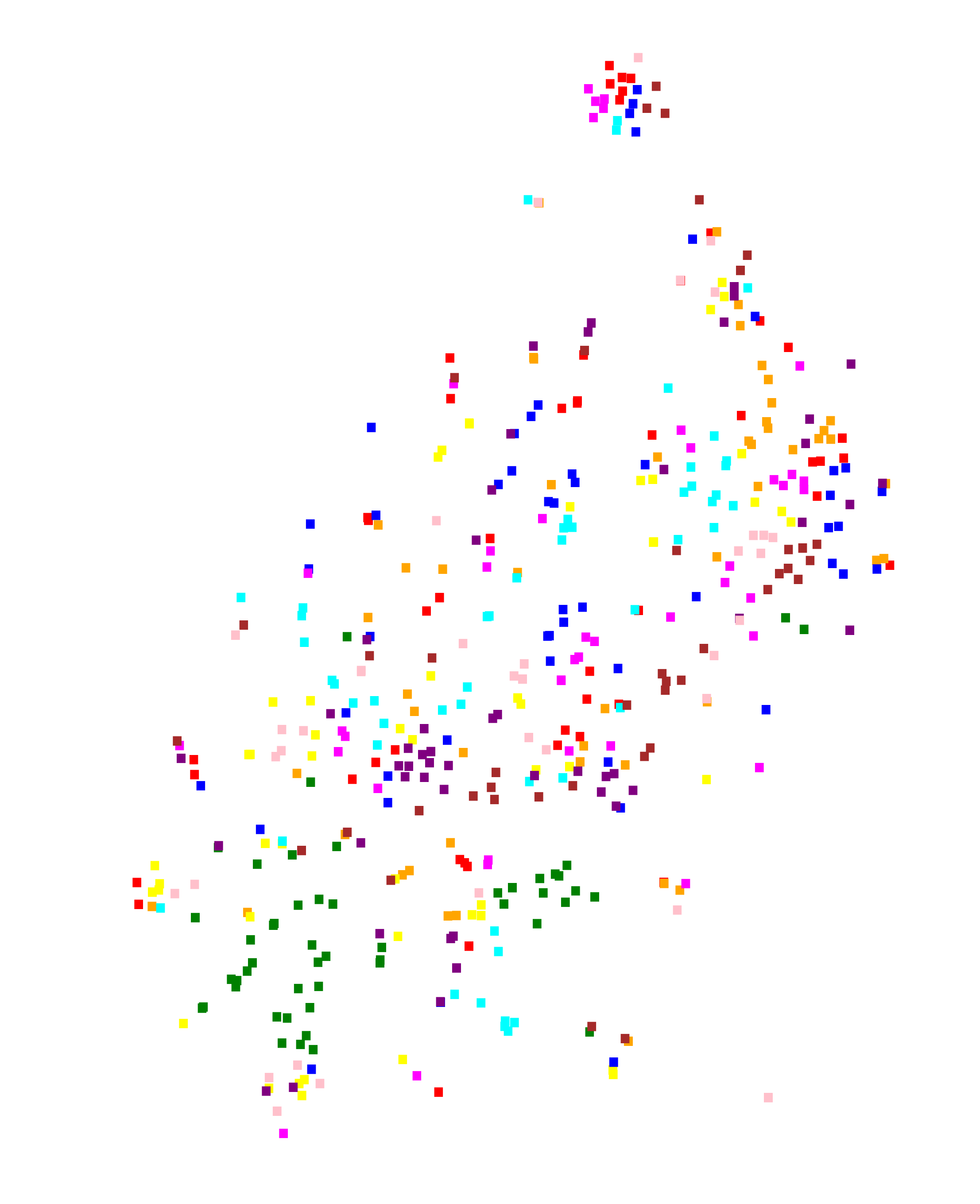}}
	\subfigure[BERT+Geo-CP\quad\quad]{
		\label{subfig-length}
		\includegraphics[width=4cm,height=4cm]{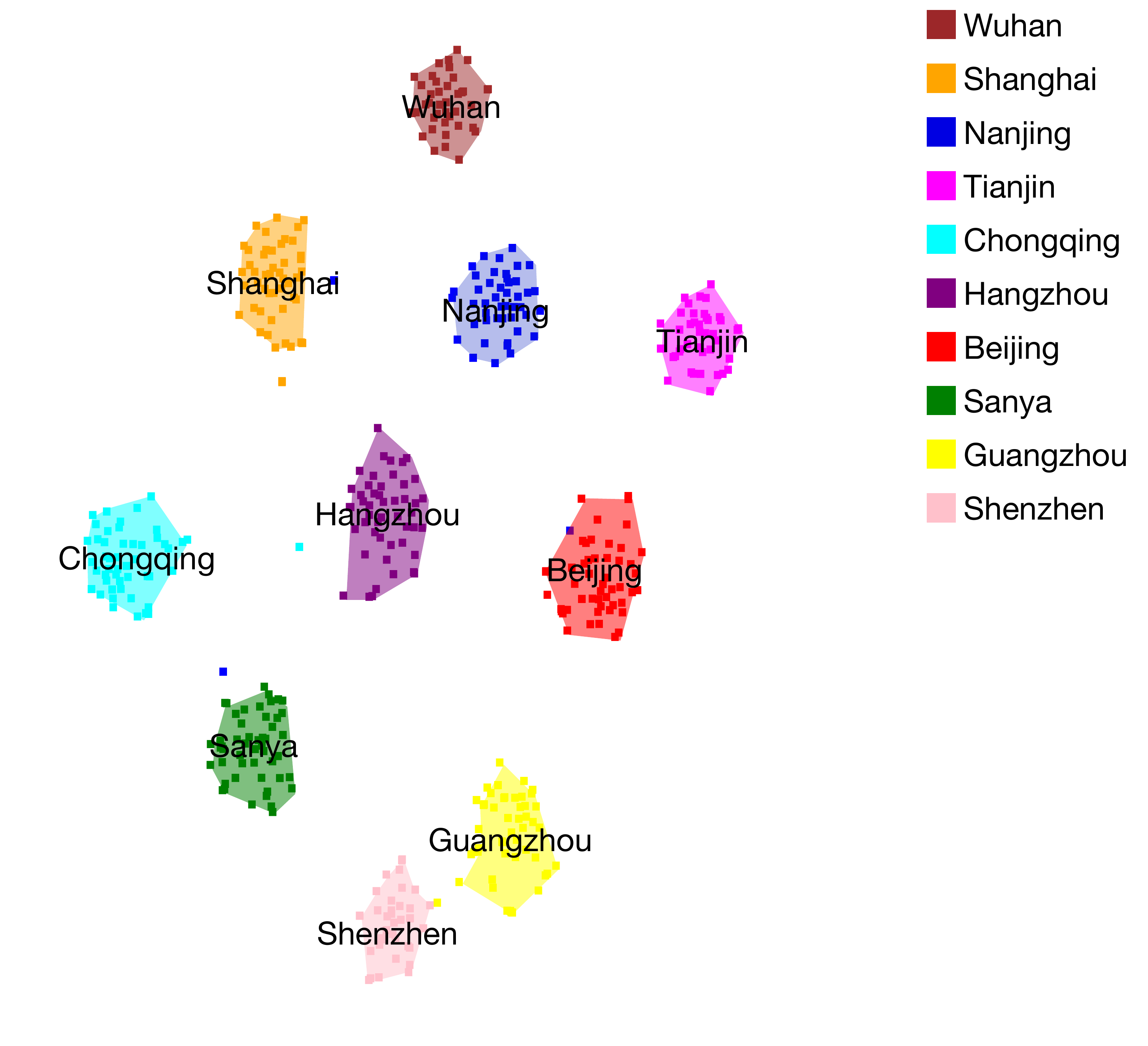}}
    \caption{The t-SNE space comparison between BERT and BERT+Geo-CP. }
    \label{fig:tsne}
\end{figure}

As shown in Figure \ref{fig:tsne}, the embeddings produced by BERT are distributed out of order in the space.
This proves that BERT does not reflect real location information of the geography query.
However, it can be observed that the embedding space of BERT+Geo-CP is more orderly, and the POIs in the same city are in the same cluster.
In addition, we also notice that the space presented by Geo-CP does have a real physical geographical location relationship.
For example, in the real world, Shanghai, Hangzhou and Nanjing are close to each other, and in the figure, the relationship among them is indeed the same.
We analyze that through Geo-CP, the language model QUERT is endowed with geographical location representation ability. 
This demonstrates the effectiveness of our proposed pre-training task.

\subsubsection{TASK 3: UCBL}
\label{subsubsec:UCBL}
To verify the effectiveness of UCBL, we compare the similarity between the click behavior related queries in the BERT and BERT+UCBL settings.
First, we select 500 queries $q$ and their click behavior related queries $q^{pos}$ from search logs.
With BERT and BERT+UCBL as encoders, we feed $q$ and $q^{pos}$ into the encoders and gain their embeddings.
For convenience, we define the embeddings of $q$ and $q^{pos}$ produced by BERT as $R_{B}^{q}$ and $R_{B}^{pos}$, and these gained from BERT+UCBL as
$R_{B+U}^{q}$ and $R_{B+U}^{pos}$.
Then we calculate cosine similarity of $(R_{B}^{q},R_{B}^{pos})$ and $(R_{B+U}^{q},R_{B+U}^{q})$.
The results are 0.7758 and 0.8278, respectively, which proves that BERT+UCBL can perceive the potential similarities of user behavior in query.

\subsubsection{TASK 4: PTOP}
\label{subsubsec:Geo-CP}
To verify the effectiveness of PTOP, we analyze the performance of different models on the order prediction subtask in QED.
To be specific, this subtask aims to judge whether the token order is transposed.
The result is reported in Table \ref{tab:ptop-order}.

Compared to BERT, StructBERT gains better performance because of its pre-training task which aims to reconstruct the token order.
However, BERT+PTOP has a performance advantage over StructBERT.
According to our analysis, PTOP directly focuses on order judgment, which is more matching with the downstream task of travel domain search.
The performance advantage of PTOP demonstrates the effectiveness of our proposed pre-training task.

\begin{table}[t]
\caption{The performance comparison of the order prediction subtask in QED.}
\begin{tabular}{llll}
\toprule
           & P     & R     & F1    \\ \midrule
BERT       & 84.77 & 77.88 & 81.18 \\
StructBERT & 81.49 & 84.33 & 82.89 \\
BERT+PTOP  & \textbf{83.48} & \textbf{84.39} & \textbf{84.20} \\
\bottomrule
\end{tabular}
\label{tab:ptop-order}
\end{table}

\begin{figure*}[htb]
    \includegraphics[scale=0.17]{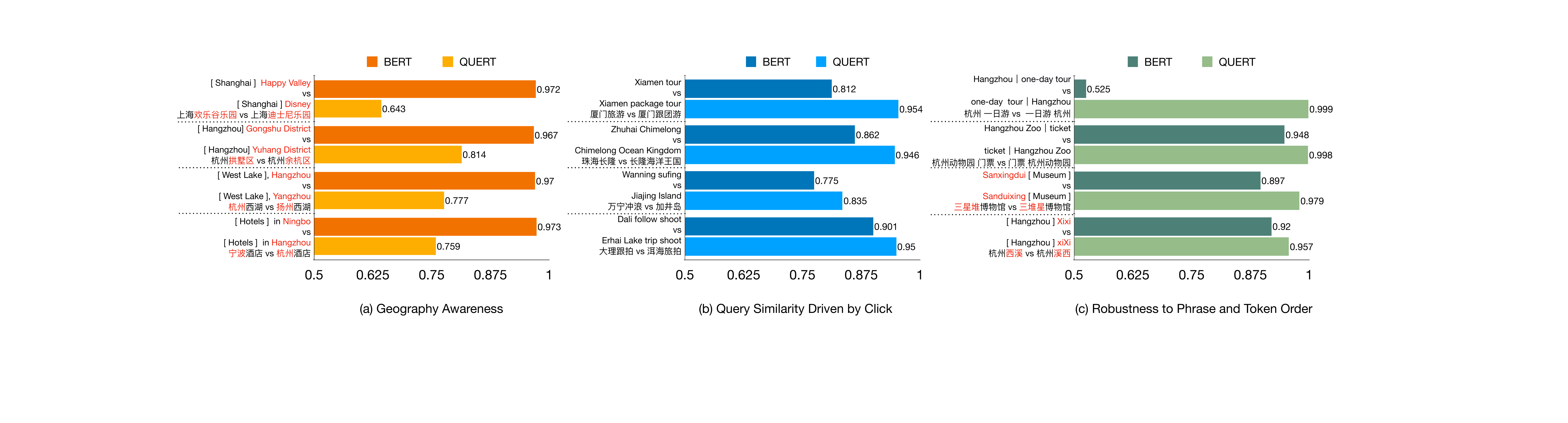}
    \caption{The comparison of cosine similarity. We compare the cosine similarity of query embeddings obtained from BERT and QUERT$_{BERT}$, respectively. ``[CLS]'' token is used to represent the whole query.
 }
\label{fig:cos}
\end{figure*}

\begin{CJK}{UTF8}{gkai}
\subsection{Case Study}
\label{subsec:case-study}
According to the different characteristics of queries targeted by different tasks, we conduct the case study in this section. As shown in Figure \ref{fig:cos}, we compare the representation ability of BERT and QUERT by evaluating the cosine similarity of embedding.
 
\textbf{Geography Awareness.}
We evaluate the sensitivity of QUERT to geography information (i.e., POI and City).
In Figure \ref{fig:cos} (a), we calculate the cosine similarity between different POI in the same city.
For example, ``Happy Valley'' and ``Disney'' are both names of parks, and BERT gives them a high degree of similarity.
However, QUERT recognizes that they are two completely different POIs and distinguishes them.
Similarly, we evaluate the same POI in different cities.
Although the two queries both contain the same POI ``West Lake'', QUERT senses that the exact essential part is the city and gives a low similarity score.
These results indicate that QUERT has an awareness of geography information, thus differentiating queries for different locations.

\textbf{Query Similarity Driven by Click.}
We also evaluate the understanding of click behavior similarity in Figure \ref{fig:cos} (b).
For the examples ``Xiamen tour'' and ``Xiamen package tour'', BERT assign low similarity for them because of literal difference.
However, QUERT understands the potential user behavior similarity for the same intention ``Searching information about the tour of Xiamen'' and assigns high similarity.
Such a phenomenon confirms that QUERT is able to understand the similarity driven by user click behavior.

\textbf{Robustness to Phrase and Token Order.}
In order to test whether QUERT is robust to phrase and token order, we choose four permutation queries as experimental objectives in Figure \ref{fig:cos} (c).
For the example of ``Hangzhou one-day tour'' and ``one-day tour Hangzhou'', QUERT has a higher tolerance for phrase permutation.
As for token permutation, QUERT identifies "Sanxingdui" as the correct form of "Sanduixing" and assigns high similarity.
These cases show that QUERT is truly robust to phrase and token order.
\end{CJK}

\vspace{-1em}
\subsection{Online Application}
\label{subsec:application}
We perform online A/B testing on Fliggy APP to validate QUERT's capabilities in the real business scenario. 
To be specific, given an unparsed query (e.g., misinput or emerging query), we gain its embedding $\mathcal{R}$ by feeding it to an encoder.
Then we calculate the cosine similarity between $\mathcal{R}$ and other embeddings of parsed queries in the database.
We select the top 20 queries with the highest similarity scores as similar queries and get the search items results corresponding to these similar queries.
And these results are considered as the potential recommendation for the unparsed query.

In our online A/B testing, we compare two buckets, each containing 10\% randomly-selected users.
We use BERT as the encoder for one bucket, and for the other, we adopt QUERT.
To measure the actual change in online business, we use two well-known metrics:  \textbf{Unique Click-Through Rate (U-CTR)} and \textbf{Page Click-Through Rate (P-CTR)}.
After running 7 days, the feedback results show that U-CTR and P-CTR increase by 0.89\% and  1.03\%, respectively.
This suggests that QUERT's unparsed query representations retrieve more relevant similar queries, signifying its aptness for embedding-based retrieval in travel domain search.


\begin{figure}[t]
\centering
	\subfigtopskip=2pt 
	\subfigure[BERT]{
		\label{subfig-type}
		\includegraphics[width=3cm,height=6cm]{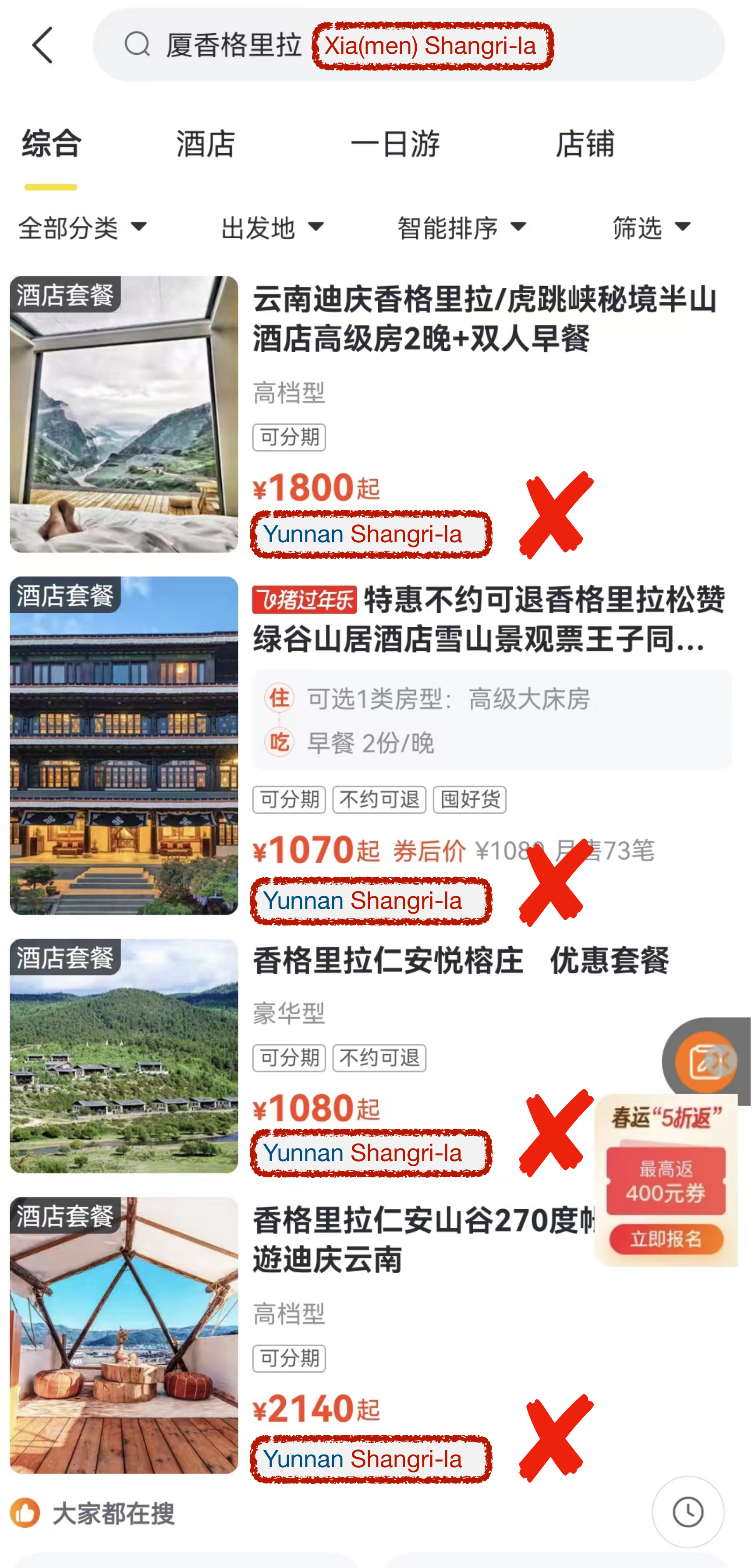}}
  \quad \quad \quad 
	\subfigure[QUERT]{
		\label{subfig-length}
		\includegraphics[width=3cm,height=6cm]{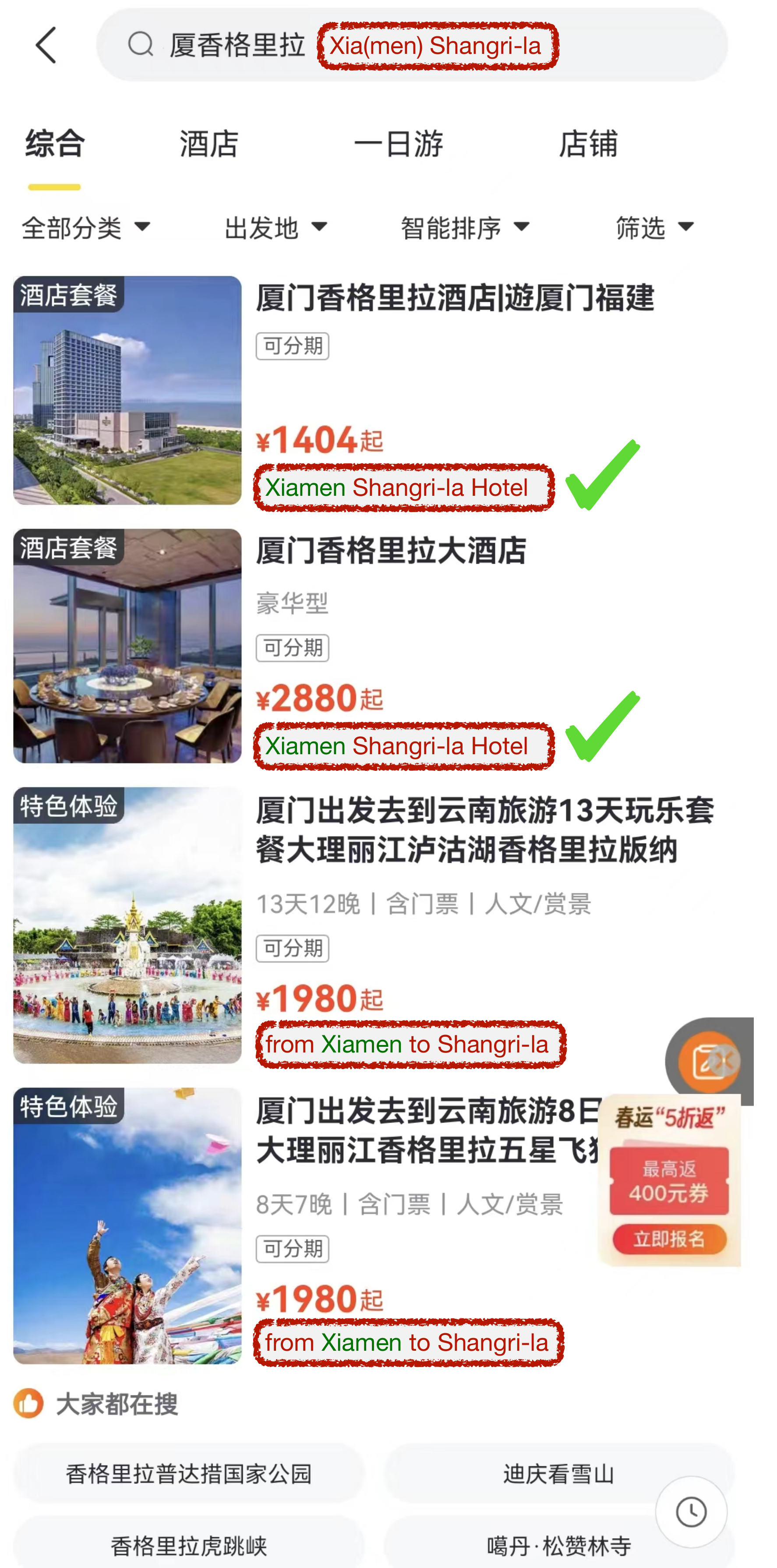}}
    \caption{Comparison of online application between BERT and QUERT.}
    \label{fig:app}
\end{figure}

In Figure \ref{fig:app}, we provide two cases. 
The query ``Xia Shangri-La'' is unparsed because it misses a token ``men''. 
And the real intention of this query is ``Searching for the hotel named  Shangri-La located in Xiamen''. 
As shown in Figure \ref{fig:app} (a), with BERT, the system mistakenly recalls items related to ``Yunnan'' because the token ``Shangri-La'' also refers to a city located in Yunnan.
Due to missing ``men'', BERT fails to capture the potential constraint ``Xiamen''.
However, with QUERT, the system recalls the right items mentioning ``Shangri-La Hotel in Xiamen'' and high-related items ``Trip from Xiamen to Shangri-La'', which proves the effectiveness of QUERT.

%% file: 060conclusion.tex
In this paper, we focus on the continual pre-training for query understanding in travel domain search.
We analyze the causes of query representation difficulties and propose a solution: QUERT, a continual pre-trained language model.
To be specific, we propose four tailored pre-training tasks: Geography-aware Mask Prediction, Geohash Code Prediction, User Click Behavior Learning, and Phrase and Token Order Prediction.
We evaluate offline performance on five downstream tasks in the travel domain.
Experimental results show that compared to BERT, the performance of QUERT on downstream tasks improves by 2.02\% and 30.93\% in supervised and unsupervised settings,  respectively.
Furthermore, the online A/B testing on Fliggy APP demonstrates that U-CTR and P-CTR increase by 0.89\% and 1.03\% when applying the QUERT as the feature encoder.

As a language model, QUERT only relies on text information. 
In future work, we plan to introduce more information (e.g., images) and explore the pre-trained multimodal in travel domain search.

%% file: kdd2023.bbl

\begin{thebibliography}{32}


\ifx \showCODEN    \undefined \def \showCODEN     #1{\unskip}     \fi
\ifx \showDOI      \undefined \def \showDOI       #1{#1}\fi
\ifx \showISBNx    \undefined \def \showISBNx     #1{\unskip}     \fi
\ifx \showISBNxiii \undefined \def \showISBNxiii  #1{\unskip}     \fi
\ifx \showISSN     \undefined \def \showISSN      #1{\unskip}     \fi
\ifx \showLCCN     \undefined \def \showLCCN      #1{\unskip}     \fi
\ifx \shownote     \undefined \def \shownote      #1{#1}          \fi
\ifx \showarticletitle \undefined \def \showarticletitle #1{#1}   \fi
\ifx \showURL      \undefined \def \showURL       {\relax}        \fi
\providecommand\bibfield[2]{#2}
\providecommand\bibinfo[2]{#2}
\providecommand\natexlab[1]{#1}
\providecommand\showeprint[2][]{arXiv:#2}

\bibitem[Araci(2019)]%
        {araci2019finbert}
\bibfield{author}{\bibinfo{person}{Dogu Araci}.}
  \bibinfo{year}{2019}\natexlab{}.
\newblock \showarticletitle{Finbert: Financial sentiment analysis with
  pre-trained language models}.
\newblock \bibinfo{journal}{\emph{arXiv preprint arXiv:1908.10063}}
  (\bibinfo{year}{2019}).
\newblock


\bibitem[Brown et~al\mbox{.}(2020)]%
        {brown2020language}
\bibfield{author}{\bibinfo{person}{Tom~B Brown}, \bibinfo{person}{Benjamin
  Mann}, \bibinfo{person}{Nick Ryder}, \bibinfo{person}{Melanie Subbiah},
  \bibinfo{person}{Jared Kaplan}, \bibinfo{person}{Prafulla Dhariwal},
  \bibinfo{person}{Arvind Neelakantan}, \bibinfo{person}{Pranav Shyam},
  \bibinfo{person}{Girish Sastry}, \bibinfo{person}{Amanda Askell},
  {et~al\mbox{.}}} \bibinfo{year}{2020}\natexlab{}.
\newblock \showarticletitle{Language Models are Few-Shot Learners}.
\newblock \bibinfo{journal}{\emph{arXiv preprint arXiv:2005.14165}}
  (\bibinfo{year}{2020}).
\newblock


\bibitem[Cowan et~al\mbox{.}(2015)]%
        {cowan2015named}
\bibfield{author}{\bibinfo{person}{Brooke Cowan}, \bibinfo{person}{Sven
  Zethelius}, \bibinfo{person}{Brittany Luk}, \bibinfo{person}{Teodora Baras},
  \bibinfo{person}{Prachi Ukarde}, {and} \bibinfo{person}{Daodao Zhang}.}
  \bibinfo{year}{2015}\natexlab{}.
\newblock \showarticletitle{Named entity recognition in travel-related search
  queries}. In \bibinfo{booktitle}{\emph{Proceedings of the AAAI conference on
  artificial intelligence}}, Vol.~\bibinfo{volume}{29}.
  \bibinfo{pages}{3935--3941}.
\newblock


\bibitem[Cui et~al\mbox{.}(2021)]%
        {cui2021pre}
\bibfield{author}{\bibinfo{person}{Yiming Cui}, \bibinfo{person}{Wanxiang Che},
  \bibinfo{person}{Ting Liu}, \bibinfo{person}{Bing Qin}, {and}
  \bibinfo{person}{Ziqing Yang}.} \bibinfo{year}{2021}\natexlab{}.
\newblock \showarticletitle{Pre-training with whole word masking for chinese
  bert}.
\newblock \bibinfo{journal}{\emph{IEEE/ACM Transactions on Audio, Speech, and
  Language Processing}}  \bibinfo{volume}{29} (\bibinfo{year}{2021}),
  \bibinfo{pages}{3504--3514}.
\newblock


\bibitem[Devlin et~al\mbox{.}(2019)]%
        {devlin-etal-2019-bert}
\bibfield{author}{\bibinfo{person}{Jacob Devlin}, \bibinfo{person}{Ming-Wei
  Chang}, \bibinfo{person}{Kenton Lee}, {and} \bibinfo{person}{Kristina
  Toutanova}.} \bibinfo{year}{2019}\natexlab{}.
\newblock \showarticletitle{{BERT}: Pre-training of Deep Bidirectional
  Transformers for Language Understanding}. In
  \bibinfo{booktitle}{\emph{Proceedings of the 2019 Conference of the North
  {A}merican Chapter of the Association for Computational Linguistics: Human
  Language Technologies, Volume 1 (Long and Short Papers)}}.
  \bibinfo{publisher}{Association for Computational Linguistics},
  \bibinfo{address}{Minneapolis, Minnesota}, \bibinfo{pages}{4171--4186}.
\newblock
\urldef\tempurl%
\url{https://doi.org/10.18653/v1/N19-1423}
\showDOI{\tempurl}


\bibitem[Dong et~al\mbox{.}(2019)]%
        {dong2019unified}
\bibfield{author}{\bibinfo{person}{Li Dong}, \bibinfo{person}{Nan Yang},
  \bibinfo{person}{Wenhui Wang}, \bibinfo{person}{Furu Wei},
  \bibinfo{person}{Xiaodong Liu}, \bibinfo{person}{Yu Wang},
  \bibinfo{person}{Jianfeng Gao}, \bibinfo{person}{Ming Zhou}, {and}
  \bibinfo{person}{Hsiao-Wuen Hon}.} \bibinfo{year}{2019}\natexlab{}.
\newblock \showarticletitle{Unified language model pre-training for natural
  language understanding and generation}.
\newblock \bibinfo{journal}{\emph{Advances in Neural Information Processing
  Systems}}  \bibinfo{volume}{32} (\bibinfo{year}{2019}).
\newblock


\bibitem[Garg and Ramakrishnan(2020)]%
        {garg2020bae}
\bibfield{author}{\bibinfo{person}{Siddhant Garg} {and}
  \bibinfo{person}{Goutham Ramakrishnan}.} \bibinfo{year}{2020}\natexlab{}.
\newblock \showarticletitle{BAE: BERT-based Adversarial Examples for Text
  Classification}. In \bibinfo{booktitle}{\emph{Proceedings of the 2020
  Conference on Empirical Methods in Natural Language Processing (EMNLP)}}.
  \bibinfo{pages}{6174--6181}.
\newblock


\bibitem[Gong et~al\mbox{.}(2022)]%
        {gong2022continual}
\bibfield{author}{\bibinfo{person}{Zheng Gong}, \bibinfo{person}{Kun Zhou},
  \bibinfo{person}{Wayne~Xin Zhao}, \bibinfo{person}{Jing Sha},
  \bibinfo{person}{Shijin Wang}, {and} \bibinfo{person}{Ji-Rong Wen}.}
  \bibinfo{year}{2022}\natexlab{}.
\newblock \showarticletitle{Continual Pre-training of Language Models for Math
  Problem Understanding with Syntax-Aware Memory Network}. In
  \bibinfo{booktitle}{\emph{Proceedings of the 60th Annual Meeting of the
  Association for Computational Linguistics (Volume 1: Long Papers)}}.
  \bibinfo{pages}{5923--5933}.
\newblock


\bibitem[Gururangan et~al\mbox{.}(2020)]%
        {gururangan2020don}
\bibfield{author}{\bibinfo{person}{Suchin Gururangan}, \bibinfo{person}{Ana
  Marasovi{\'c}}, \bibinfo{person}{Swabha Swayamdipta}, \bibinfo{person}{Kyle
  Lo}, \bibinfo{person}{Iz Beltagy}, \bibinfo{person}{Doug Downey}, {and}
  \bibinfo{person}{Noah~A Smith}.} \bibinfo{year}{2020}\natexlab{}.
\newblock \showarticletitle{Don’t Stop Pretraining: Adapt Language Models to
  Domains and Tasks}. In \bibinfo{booktitle}{\emph{Proceedings of the 58th
  Annual Meeting of the Association for Computational Linguistics}}.
  \bibinfo{pages}{8342--8360}.
\newblock


\bibitem[Huang et~al\mbox{.}(2022)]%
        {huang2022ernie}
\bibfield{author}{\bibinfo{person}{Jizhou Huang}, \bibinfo{person}{Haifeng
  Wang}, \bibinfo{person}{Yibo Sun}, \bibinfo{person}{Yunsheng Shi},
  \bibinfo{person}{Zhengjie Huang}, \bibinfo{person}{An Zhuo}, {and}
  \bibinfo{person}{Shikun Feng}.} \bibinfo{year}{2022}\natexlab{}.
\newblock \showarticletitle{ERNIE-GeoL: A Geography-and-Language Pre-trained
  Model and its Applications in Baidu Maps}. In
  \bibinfo{booktitle}{\emph{Proceedings of the 28th ACM SIGKDD Conference on
  Knowledge Discovery and Data Mining}}. \bibinfo{pages}{3029--3039}.
\newblock


\bibitem[Joshi et~al\mbox{.}(2020)]%
        {joshi2020spanbert}
\bibfield{author}{\bibinfo{person}{Mandar Joshi}, \bibinfo{person}{Danqi Chen},
  \bibinfo{person}{Yinhan Liu}, \bibinfo{person}{Daniel~S Weld},
  \bibinfo{person}{Luke Zettlemoyer}, {and} \bibinfo{person}{Omer Levy}.}
  \bibinfo{year}{2020}\natexlab{}.
\newblock \showarticletitle{Spanbert: Improving pre-training by representing
  and predicting spans}.
\newblock \bibinfo{journal}{\emph{Transactions of the Association for
  Computational Linguistics}}  \bibinfo{volume}{8} (\bibinfo{year}{2020}),
  \bibinfo{pages}{64--77}.
\newblock


\bibitem[Lan et~al\mbox{.}(2019)]%
        {lan2019albert}
\bibfield{author}{\bibinfo{person}{Zhenzhong Lan}, \bibinfo{person}{Mingda
  Chen}, \bibinfo{person}{Sebastian Goodman}, \bibinfo{person}{Kevin Gimpel},
  \bibinfo{person}{Piyush Sharma}, {and} \bibinfo{person}{Radu Soricut}.}
  \bibinfo{year}{2019}\natexlab{}.
\newblock \showarticletitle{ALBERT: A Lite BERT for Self-supervised Learning of
  Language Representations}. In \bibinfo{booktitle}{\emph{International
  Conference on Learning Representations}}.
\newblock


\bibitem[Lee et~al\mbox{.}(2020)]%
        {lee2020biobert}
\bibfield{author}{\bibinfo{person}{Jinhyuk Lee}, \bibinfo{person}{Wonjin Yoon},
  \bibinfo{person}{Sungdong Kim}, \bibinfo{person}{Donghyeon Kim},
  \bibinfo{person}{Sunkyu Kim}, \bibinfo{person}{Chan~Ho So}, {and}
  \bibinfo{person}{Jaewoo Kang}.} \bibinfo{year}{2020}\natexlab{}.
\newblock \showarticletitle{BioBERT: a pre-trained biomedical language
  representation model for biomedical text mining}.
\newblock \bibinfo{journal}{\emph{Bioinformatics}} \bibinfo{volume}{36},
  \bibinfo{number}{4} (\bibinfo{year}{2020}), \bibinfo{pages}{1234--1240}.
\newblock


\bibitem[Lewis et~al\mbox{.}(2020)]%
        {lewis-etal-2020-bart}
\bibfield{author}{\bibinfo{person}{Mike Lewis}, \bibinfo{person}{Yinhan Liu},
  \bibinfo{person}{Naman Goyal}, \bibinfo{person}{Marjan Ghazvininejad},
  \bibinfo{person}{Abdelrahman Mohamed}, \bibinfo{person}{Omer Levy},
  \bibinfo{person}{Veselin Stoyanov}, {and} \bibinfo{person}{Luke
  Zettlemoyer}.} \bibinfo{year}{2020}\natexlab{}.
\newblock \showarticletitle{{BART}: Denoising Sequence-to-Sequence Pre-training
  for Natural Language Generation, Translation, and Comprehension}. In
  \bibinfo{booktitle}{\emph{Proceedings of the 58th Annual Meeting of the
  Association for Computational Linguistics}}. \bibinfo{publisher}{Association
  for Computational Linguistics}, \bibinfo{address}{Online},
  \bibinfo{pages}{7871--7880}.
\newblock
\urldef\tempurl%
\url{https://doi.org/10.18653/v1/2020.acl-main.703}
\showDOI{\tempurl}


\bibitem[Liang et~al\mbox{.}(2020)]%
        {liang2020bond}
\bibfield{author}{\bibinfo{person}{Chen Liang}, \bibinfo{person}{Yue Yu},
  \bibinfo{person}{Haoming Jiang}, \bibinfo{person}{Siawpeng Er},
  \bibinfo{person}{Ruijia Wang}, \bibinfo{person}{Tuo Zhao}, {and}
  \bibinfo{person}{Chao Zhang}.} \bibinfo{year}{2020}\natexlab{}.
\newblock \showarticletitle{Bond: Bert-assisted open-domain named entity
  recognition with distant supervision}. In
  \bibinfo{booktitle}{\emph{Proceedings of the 26th ACM SIGKDD International
  Conference on Knowledge Discovery \& Data Mining}}.
  \bibinfo{pages}{1054--1064}.
\newblock


\bibitem[Liu et~al\mbox{.}(2021)]%
        {liu2021geo}
\bibfield{author}{\bibinfo{person}{Xiao Liu}, \bibinfo{person}{Juan Hu},
  \bibinfo{person}{Qi Shen}, {and} \bibinfo{person}{Huan Chen}.}
  \bibinfo{year}{2021}\natexlab{}.
\newblock \showarticletitle{Geo-BERT Pre-training Model for Query Rewriting in
  POI Search}. In \bibinfo{booktitle}{\emph{Findings of the Association for
  Computational Linguistics: EMNLP 2021}}. \bibinfo{pages}{2209--2214}.
\newblock


\bibitem[Liu et~al\mbox{.}(2019)]%
        {liu2019roberta}
\bibfield{author}{\bibinfo{person}{Yinhan Liu}, \bibinfo{person}{Myle Ott},
  \bibinfo{person}{Naman Goyal}, \bibinfo{person}{Jingfei Du},
  \bibinfo{person}{Mandar Joshi}, \bibinfo{person}{Danqi Chen},
  \bibinfo{person}{Omer Levy}, \bibinfo{person}{Mike Lewis},
  \bibinfo{person}{Luke Zettlemoyer}, {and} \bibinfo{person}{Veselin
  Stoyanov}.} \bibinfo{year}{2019}\natexlab{}.
\newblock \showarticletitle{Roberta: A robustly optimized bert pretraining
  approach}.
\newblock \bibinfo{journal}{\emph{arXiv preprint arXiv:1907.11692}}
  (\bibinfo{year}{2019}).
\newblock


\bibitem[Loshchilov and Hutter(2018)]%
        {loshchilov2018decoupled}
\bibfield{author}{\bibinfo{person}{Ilya Loshchilov} {and}
  \bibinfo{person}{Frank Hutter}.} \bibinfo{year}{2018}\natexlab{}.
\newblock \showarticletitle{Decoupled Weight Decay Regularization}. In
  \bibinfo{booktitle}{\emph{International Conference on Learning
  Representations}}.
\newblock


\bibitem[Ma et~al\mbox{.}(2021a)]%
        {ma2021prop}
\bibfield{author}{\bibinfo{person}{Xinyu Ma}, \bibinfo{person}{Jiafeng Guo},
  \bibinfo{person}{Ruqing Zhang}, \bibinfo{person}{Yixing Fan},
  \bibinfo{person}{Xiang Ji}, {and} \bibinfo{person}{Xueqi Cheng}.}
  \bibinfo{year}{2021}\natexlab{a}.
\newblock \showarticletitle{PROP: pre-training with representative words
  prediction for ad-hoc retrieval}. In \bibinfo{booktitle}{\emph{Proceedings of
  the 14th ACM International Conference on Web Search and Data Mining}}.
  \bibinfo{pages}{283--291}.
\newblock


\bibitem[Ma et~al\mbox{.}(2021b)]%
        {ma2021b}
\bibfield{author}{\bibinfo{person}{Xinyu Ma}, \bibinfo{person}{Jiafeng Guo},
  \bibinfo{person}{Ruqing Zhang}, \bibinfo{person}{Yixing Fan},
  \bibinfo{person}{Yingyan Li}, {and} \bibinfo{person}{Xueqi Cheng}.}
  \bibinfo{year}{2021}\natexlab{b}.
\newblock \showarticletitle{B-PROP: bootstrapped pre-training with
  representative words prediction for ad-hoc retrieval}. In
  \bibinfo{booktitle}{\emph{Proceedings of the 44th International ACM SIGIR
  Conference on Research and Development in Information Retrieval}}.
  \bibinfo{pages}{1513--1522}.
\newblock


\bibitem[M{\"u}ller et~al\mbox{.}(2020)]%
        {muller2020covid}
\bibfield{author}{\bibinfo{person}{Martin M{\"u}ller}, \bibinfo{person}{Marcel
  Salath{\'e}}, {and} \bibinfo{person}{Per~E Kummervold}.}
  \bibinfo{year}{2020}\natexlab{}.
\newblock \showarticletitle{Covid-twitter-bert: A natural language processing
  model to analyse covid-19 content on twitter}.
\newblock \bibinfo{journal}{\emph{arXiv preprint arXiv:2005.07503}}
  (\bibinfo{year}{2020}).
\newblock


\bibitem[Nguyen et~al\mbox{.}(2020)]%
        {nguyen-etal-2020-bertweet}
\bibfield{author}{\bibinfo{person}{Dat~Quoc Nguyen}, \bibinfo{person}{Thanh
  Vu}, {and} \bibinfo{person}{Anh Tuan~Nguyen}.}
  \bibinfo{year}{2020}\natexlab{}.
\newblock \showarticletitle{{BERT}weet: A pre-trained language model for
  {E}nglish Tweets}. In \bibinfo{booktitle}{\emph{Proceedings of the 2020
  Conference on Empirical Methods in Natural Language Processing: System
  Demonstrations}}. \bibinfo{publisher}{Association for Computational
  Linguistics}, \bibinfo{address}{Online}, \bibinfo{pages}{9--14}.
\newblock
\urldef\tempurl%
\url{https://doi.org/10.18653/v1/2020.emnlp-demos.2}
\showDOI{\tempurl}


\bibitem[Paszke et~al\mbox{.}(2019)]%
        {NEURIPS2019_bdbca288}
\bibfield{author}{\bibinfo{person}{Adam Paszke}, \bibinfo{person}{Sam Gross},
  \bibinfo{person}{Francisco Massa}, \bibinfo{person}{Adam Lerer},
  \bibinfo{person}{James Bradbury}, \bibinfo{person}{Gregory Chanan},
  \bibinfo{person}{Trevor Killeen}, \bibinfo{person}{Zeming Lin},
  \bibinfo{person}{Natalia Gimelshein}, \bibinfo{person}{Luca Antiga},
  \bibinfo{person}{Alban Desmaison}, \bibinfo{person}{Andreas Kopf},
  \bibinfo{person}{Edward Yang}, \bibinfo{person}{Zachary DeVito},
  \bibinfo{person}{Martin Raison}, \bibinfo{person}{Alykhan Tejani},
  \bibinfo{person}{Sasank Chilamkurthy}, \bibinfo{person}{Benoit Steiner},
  \bibinfo{person}{Lu Fang}, \bibinfo{person}{Junjie Bai}, {and}
  \bibinfo{person}{Soumith Chintala}.} \bibinfo{year}{2019}\natexlab{}.
\newblock \showarticletitle{PyTorch: An Imperative Style, High-Performance Deep
  Learning Library}. In \bibinfo{booktitle}{\emph{Advances in Neural
  Information Processing Systems}},
  \bibfield{editor}{\bibinfo{person}{H.~Wallach},
  \bibinfo{person}{H.~Larochelle}, \bibinfo{person}{A.~Beygelzimer},
  \bibinfo{person}{F.~d\textquotesingle Alch\'{e}-Buc},
  \bibinfo{person}{E.~Fox}, {and} \bibinfo{person}{R.~Garnett}} (Eds.),
  Vol.~\bibinfo{volume}{32}. \bibinfo{publisher}{Curran Associates, Inc.}
\newblock
\urldef\tempurl%
\url{https://proceedings.neurips.cc/paper/2019/file/bdbca288fee7f92f2bfa9f7012727740-Paper.pdf}
\showURL{%
\tempurl}


\bibitem[Sun et~al\mbox{.}(2019)]%
        {sun2019ernie}
\bibfield{author}{\bibinfo{person}{Yu Sun}, \bibinfo{person}{Shuohuan Wang},
  \bibinfo{person}{Yukun Li}, \bibinfo{person}{Shikun Feng},
  \bibinfo{person}{Xuyi Chen}, \bibinfo{person}{Han Zhang},
  \bibinfo{person}{Xin Tian}, \bibinfo{person}{Danxiang Zhu},
  \bibinfo{person}{Hao Tian}, {and} \bibinfo{person}{Hua Wu}.}
  \bibinfo{year}{2019}\natexlab{}.
\newblock \showarticletitle{ERNIE: Enhanced Representation through Knowledge
  Integration}.
\newblock \bibinfo{journal}{\emph{arXiv e-prints}} (\bibinfo{year}{2019}),
  \bibinfo{pages}{arXiv--1904}.
\newblock


\bibitem[Van~der Maaten and Hinton(2008)]%
        {van2008visualizing}
\bibfield{author}{\bibinfo{person}{Laurens Van~der Maaten} {and}
  \bibinfo{person}{Geoffrey Hinton}.} \bibinfo{year}{2008}\natexlab{}.
\newblock \showarticletitle{Visualizing data using t-SNE.}
\newblock \bibinfo{journal}{\emph{Journal of machine learning research}}
  \bibinfo{volume}{9}, \bibinfo{number}{11} (\bibinfo{year}{2008}).
\newblock


\bibitem[Vaswani et~al\mbox{.}(2017)]%
        {vaswani2017attention}
\bibfield{author}{\bibinfo{person}{Ashish Vaswani}, \bibinfo{person}{Noam
  Shazeer}, \bibinfo{person}{Niki Parmar}, \bibinfo{person}{Jakob Uszkoreit},
  \bibinfo{person}{Llion Jones}, \bibinfo{person}{Aidan~N Gomez},
  \bibinfo{person}{{\L}ukasz Kaiser}, {and} \bibinfo{person}{Illia
  Polosukhin}.} \bibinfo{year}{2017}\natexlab{}.
\newblock \showarticletitle{Attention is all you need}.
\newblock \bibinfo{journal}{\emph{Advances in neural information processing
  systems}}  \bibinfo{volume}{30} (\bibinfo{year}{2017}).
\newblock


\bibitem[Wang et~al\mbox{.}(2019)]%
        {wang2019structbert}
\bibfield{author}{\bibinfo{person}{Wei Wang}, \bibinfo{person}{Bin Bi},
  \bibinfo{person}{Ming Yan}, \bibinfo{person}{Chen Wu},
  \bibinfo{person}{Jiangnan Xia}, \bibinfo{person}{Zuyi Bao},
  \bibinfo{person}{Liwei Peng}, {and} \bibinfo{person}{Luo Si}.}
  \bibinfo{year}{2019}\natexlab{}.
\newblock \showarticletitle{StructBERT: Incorporating Language Structures into
  Pre-training for Deep Language Understanding}. In
  \bibinfo{booktitle}{\emph{International Conference on Learning
  Representations}}.
\newblock


\bibitem[Wolf et~al\mbox{.}(2020)]%
        {wolf-etal-2020-transformers}
\bibfield{author}{\bibinfo{person}{Thomas Wolf}, \bibinfo{person}{Lysandre
  Debut}, \bibinfo{person}{Victor Sanh}, \bibinfo{person}{Julien Chaumond},
  \bibinfo{person}{Clement Delangue}, \bibinfo{person}{Anthony Moi},
  \bibinfo{person}{Pierric Cistac}, \bibinfo{person}{Tim Rault},
  \bibinfo{person}{Remi Louf}, \bibinfo{person}{Morgan Funtowicz},
  \bibinfo{person}{Joe Davison}, \bibinfo{person}{Sam Shleifer},
  \bibinfo{person}{Patrick von Platen}, \bibinfo{person}{Clara Ma},
  \bibinfo{person}{Yacine Jernite}, \bibinfo{person}{Julien Plu},
  \bibinfo{person}{Canwen Xu}, \bibinfo{person}{Teven Le~Scao},
  \bibinfo{person}{Sylvain Gugger}, \bibinfo{person}{Mariama Drame},
  \bibinfo{person}{Quentin Lhoest}, {and} \bibinfo{person}{Alexander Rush}.}
  \bibinfo{year}{2020}\natexlab{}.
\newblock \showarticletitle{Transformers: State-of-the-Art Natural Language
  Processing}. In \bibinfo{booktitle}{\emph{Proceedings of the 2020 Conference
  on Empirical Methods in Natural Language Processing: System Demonstrations}}.
  \bibinfo{publisher}{Association for Computational Linguistics},
  \bibinfo{address}{Online}, \bibinfo{pages}{38--45}.
\newblock
\urldef\tempurl%
\url{https://doi.org/10.18653/v1/2020.emnlp-demos.6}
\showDOI{\tempurl}


\bibitem[Xu et~al\mbox{.}(2022)]%
        {xu2022g2net}
\bibfield{author}{\bibinfo{person}{Jia Xu}, \bibinfo{person}{Fei Xiong},
  \bibinfo{person}{Zulong Chen}, \bibinfo{person}{Mingyuan Tao},
  \bibinfo{person}{Liangyue Li}, {and} \bibinfo{person}{Quan Lu}.}
  \bibinfo{year}{2022}\natexlab{}.
\newblock \showarticletitle{G2NET: A General Geography-Aware Representation
  Network for Hotel Search Ranking}. In \bibinfo{booktitle}{\emph{Proceedings
  of the 28th ACM SIGKDD Conference on Knowledge Discovery and Data Mining}}.
  \bibinfo{pages}{4237--4247}.
\newblock


\bibitem[Yang et~al\mbox{.}(2019)]%
        {yang2019xlnet}
\bibfield{author}{\bibinfo{person}{Zhilin Yang}, \bibinfo{person}{Zihang Dai},
  \bibinfo{person}{Yiming Yang}, \bibinfo{person}{Jaime Carbonell},
  \bibinfo{person}{Russ~R Salakhutdinov}, {and} \bibinfo{person}{Quoc~V Le}.}
  \bibinfo{year}{2019}\natexlab{}.
\newblock \showarticletitle{Xlnet: Generalized autoregressive pretraining for
  language understanding}.
\newblock \bibinfo{journal}{\emph{Advances in neural information processing
  systems}}  \bibinfo{volume}{32} (\bibinfo{year}{2019}).
\newblock


\bibitem[Zhang et~al\mbox{.}(2019)]%
        {zhang2019ernie}
\bibfield{author}{\bibinfo{person}{Zhengyan Zhang}, \bibinfo{person}{Xu Han},
  \bibinfo{person}{Zhiyuan Liu}, \bibinfo{person}{Xin Jiang},
  \bibinfo{person}{Maosong Sun}, {and} \bibinfo{person}{Qun Liu}.}
  \bibinfo{year}{2019}\natexlab{}.
\newblock \showarticletitle{ERNIE: Enhanced Language Representation with
  Informative Entities}. In \bibinfo{booktitle}{\emph{Proceedings of the 57th
  Annual Meeting of the Association for Computational Linguistics}}.
  \bibinfo{pages}{1441--1451}.
\newblock


\bibitem[Zhang et~al\mbox{.}(2021)]%
        {zhang2021mengzi}
\bibfield{author}{\bibinfo{person}{Zhuosheng Zhang}, \bibinfo{person}{Hanqing
  Zhang}, \bibinfo{person}{Keming Chen}, \bibinfo{person}{Yuhang Guo},
  \bibinfo{person}{Jingyun Hua}, \bibinfo{person}{Yulong Wang}, {and}
  \bibinfo{person}{Ming Zhou}.} \bibinfo{year}{2021}\natexlab{}.
\newblock \showarticletitle{Mengzi: Towards lightweight yet ingenious
  pre-trained models for chinese}.
\newblock \bibinfo{journal}{\emph{arXiv preprint arXiv:2110.06696}}
  (\bibinfo{year}{2021}).
\newblock


\end{thebibliography}
